\documentclass[10pt,twocolumn,letterpaper]{article}

\usepackage[pagenumbers]{iccv} 

\usepackage{amsmath}
\usepackage{amssymb}
\usepackage{mathtools}
\usepackage{amsthm}
\usepackage{bm}
\usepackage{xcolor}

\definecolor{iccvblue}{rgb}{0.21,0.49,0.74}
\usepackage[pagebackref,breaklinks,colorlinks,allcolors=iccvblue]{hyperref}
\usepackage{amssymb}
\usepackage{amsmath}       
\usepackage{mathtools}
\usepackage{amsthm}
\usepackage{amsfonts}       
\usepackage{nicefrac}       
\usepackage{tcolorbox}
\usepackage{microtype}
\usepackage{url}
\usepackage{graphicx}
\usepackage{pifont}
\usepackage{booktabs} 
\usepackage{xcolor}        
\usepackage{colortbl}
\usepackage{makecell}
\usepackage{array}
\usepackage{subcaption}
\usepackage{diagbox}
\usepackage{diagbox}
\usepackage{booktabs}  
\usepackage[capitalize]{cleveref}
\usepackage{multirow}  
\usepackage[dvipsnames]{xcolor} 
\usepackage[ruled,vlined,linesnumbered]{algorithm2e}
\usepackage{algorithm2e}

\usepackage[accsupp]{axessibility}
\usepackage{bbding}

\SetKwInput{KwInput}{\textcolor{blue}{Input}}    
\SetKwInput{KwOutput}{\textcolor{blue}{Output}}  
\SetCommentSty{itshape\color{gray}}              
\SetKw{KwTo}{to}                                 
\SetKw{KwRet}{\textcolor{blue}{Return}}          

\def\paperID{2729} 
\def\confName{ICCV}
\def\confYear{2025}

\title{
  \raisebox{-0.8ex}{%
    \includegraphics[
      width=0.8cm,
      height=0.80cm,
      keepaspectratio
    ]{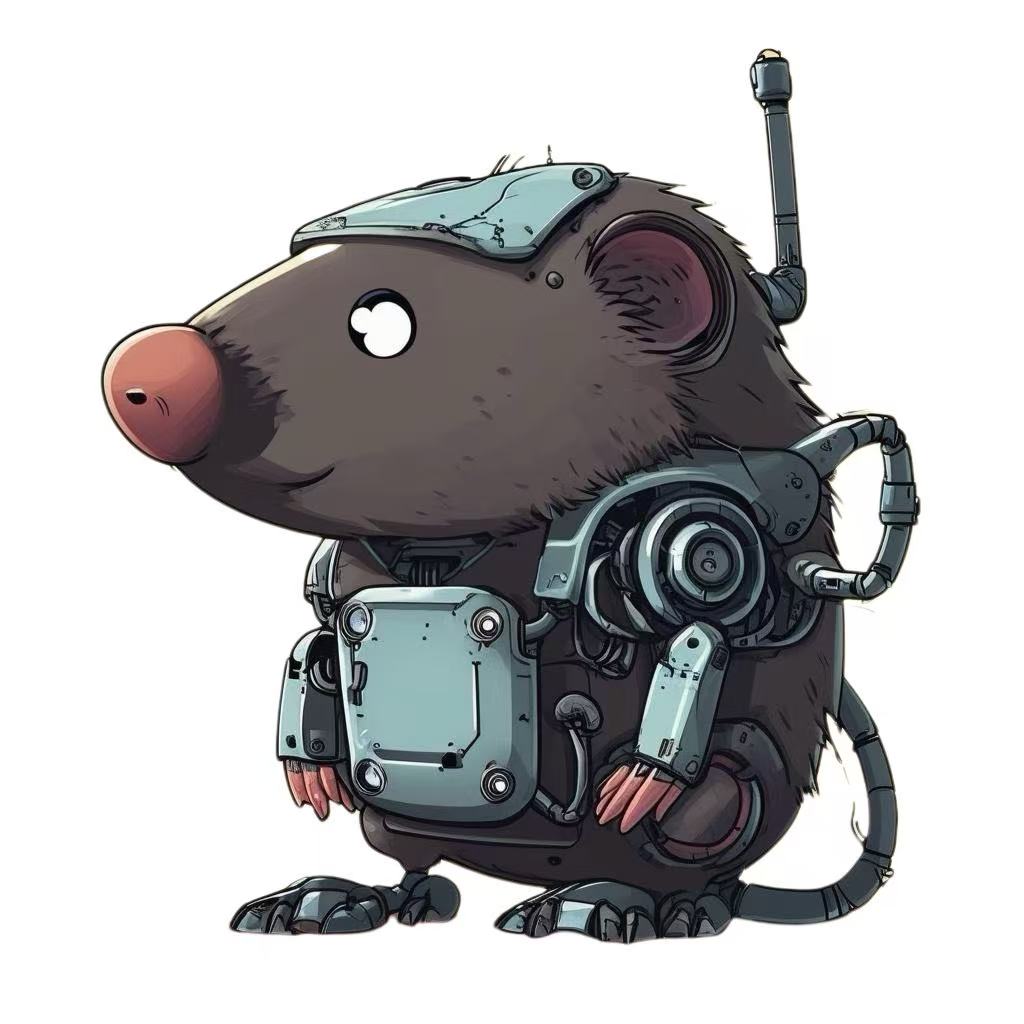}
  }
    MoLe-VLA: Dynamic Layer-skipping Vision Language Action Model via Mixture-of-Layers for Efficient Robot Manipulation}

\author{
Rongyu Zhang\textsuperscript{\rm 1,2,3,4$^{*}$}, Menghang Dong\textsuperscript{\rm 3$^{*}$}, Yuan Zhang\textsuperscript{\rm 3}, Liang Heng\textsuperscript{\rm 3}, Xiaowei Chi\textsuperscript{\rm 5}, \\Gaole Dai\textsuperscript{\rm 3,4},
Li Du\textsuperscript{\rm 1}, Yuan Du\textsuperscript{\rm 1}~\textsuperscript{\Envelope}, Shanghang Zhang\textsuperscript{\rm 3}~\textsuperscript{\Envelope}
\vspace{0.2cm}\\
\textsuperscript{\rm 1}Nanjing University; \textsuperscript{\rm 2}The Hong Kong Polytechnic University;\\
\textsuperscript{\rm 3}State Key Laboratory of Multimedia Information Processing, School of Computer Science,\\Peking University;  \textsuperscript{\rm 4}Beijing Academy of Artificial Intelligence; \\ \textsuperscript{\rm 5}The Hong Kong University of Science and Technology\\
$^{*}$ Equal contribution, \Envelope \ Corresponding author,
\textbf{Project web page:} \href{https://sites.google.com/view/mole-vla}{MoLe-VLA-Web}
}

\begin{document}
\twocolumn[
{%
\renewcommand\twocolumn[1][]{#1}
\maketitle
\begin{center}
\centering
\begin{minipage}[t]{\linewidth}
\includegraphics[width=\textwidth]{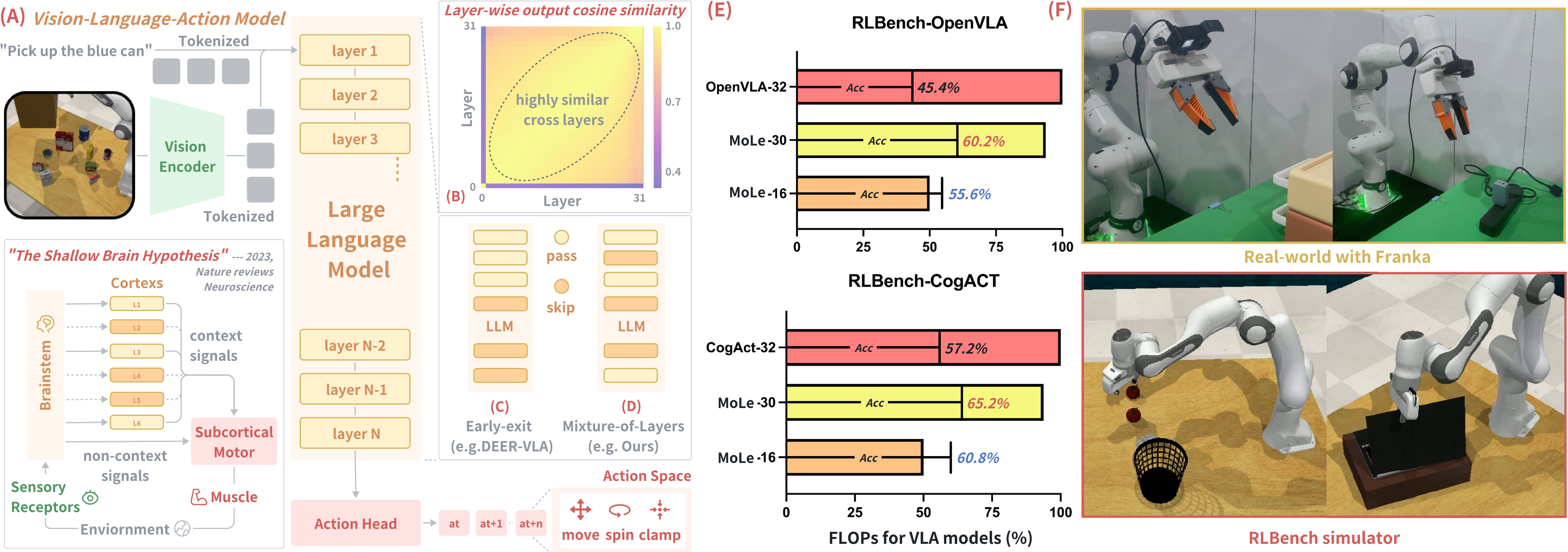}
\vspace{-0.5cm}
{\captionsetup{hypcap=false}  
\captionof{figure}{\footnotesize{
Overview of our proposed \textbf{\textit{MoLe-VLA}}: Our proposed framework integrates dynamic layer activation, a novel Spatial-Temporal Aware Router (STAR), and self-knowledge distillation (CogKD) to achieve efficient and adaptive performance in robotic applications. MoLe reduces computational costs while enhancing model performance, enabling resource-constrained platforms to benefit from MLLMs.
}}
\label{fig:teaser}}
\end{minipage}
\end{center}
}]

\begin{abstract}
Multimodal Large Language Models (MLLMs) excel in understanding complex language and visual data, enabling generalist robotic systems to interpret instructions and perform embodied tasks. Nevertheless, their real-world deployment is hindered by substantial computational and storage demands. Recent insights into the homogeneous patterns in the LLM layer have inspired sparsification techniques to address these challenges, such as early exit and token pruning. However, these methods often neglect the critical role of the final layers that encode the semantic information most relevant to downstream robotic tasks. 
Aligning with the recent breakthrough of the Shallow Brain Hypothesis (SBH) in neuroscience and the mixture of experts in model sparsification, we conceptualize each LLM layer as an expert and propose a \textbf{\underline{M}}ixture-\textbf{\underline{o}}f-\textbf{\underline{L}}ay\textbf{\underline{e}}rs \textbf{\underline{V}}ision-\textbf{\underline{L}}anguage-\textbf{\underline{A}}ction model (\textbf{\textit{MoLe-VLA}} or simply \textbf{\textit{MoLe}}) architecture for dynamic LLM layer activation. We introduce a Spatial-Temporal Aware Router (STAR) for MoLe to selectively activate only parts of the layers based on the robot’s current state, miming the brain's distinct signal pathways specialized for cognition and causal reasoning.
Additionally, to compensate for the cognition ability of LLM lost in MoLe, we devise a cognition self-knowledge distillation (CogKD) to enhance the understanding of task demands and generate task-relevant action sequences by leveraging cognition features. Extensive experiments in both RLBench simulation and real-world environments demonstrate the superiority of MoLe-VLA in both efficiency and performance, achieving performance improvement of 8\% mean success rate across ten tasks while reducing at most $\times 5.6$ computational costs in LLM.
\end{abstract}    
\section{Introduction}
\label{sec:intro}

The rapid advancements in multimodal large language models (MLLMs)~\cite{luo2024llm, an2024mc, li2022blip, alayrac2022flamingo, radford2021learning} have demonstrated their ability to integrate complex language and visual representations, inspiring the development of generalist robots and embodied agents capable of vision-language comprehension, human interaction, and flexible problem-solving in manipulation tasks. Preliminary vision language action (VLA) models~\cite{liu2024robomambaefficientvisionlanguageactionmodel, kim2024openvlaopensourcevisionlanguageactionmodel, li2023vision, li2024cogactfoundationalvisionlanguageactionmodel}, such as RT-2~\cite{brohan2023rt} and OpenVLA~\cite{kim2024openvlaopensourcevisionlanguageactionmodel}, have shown the feasibility of using MLLMs for end-to-end robotic control, enabling robust policies and emergent abilities, including generalization to unseen objects and understanding novel commands. 
However, deploying MLLMs in real-world robotic systems faces significant challenges due to their high computational demands, including substantial memory usage, power consumption, and time delays, which conflict with robotic platform resource-constrained and real-time requirements. For example, a 7B VLA model running on a commercial-grade GPU like the RTX 4090 generally achieves an inference frequency of approximately $5-12$ Hz, which falls significantly short of the $50-1000$ Hz control frequency required by the Franka robotic arm. 

Recent studies~\cite{yue2024deervladynamicinferencemultimodal, raposo2024mixtureofdepthsdynamicallyallocatingcompute} have uncovered significant redundancy in LLM layer, particularly in robotic tasks, where homogeneous patterns across layers lead to high computational costs with limited performance gains. For instance, DeeR~\cite{yue2024deervladynamicinferencemultimodal} demonstrated that using all 24 layers of the Flamingo~\cite{li2023vision} model improves task success rates by only $3.2\%$ compared to using six layers, while computational costs increase 4x on the Calvin LH-MTLC~\cite{mees2022calvin}. Similarly, our analysis of OpenVLA~\cite{kim2024openvlaopensourcevisionlanguageactionmodel} with RLBench~\cite{james2020rlbench} in \cref{fig:teaser} (A) reveals that cosine similarity between consecutive layer outputs exceeds \textit{90\%}, while features from the first and last layers differ significantly. This suggests the potential for skipping adjacent layers to reduce computation but also highlights the limitations of early-exit strategies~\cite{yue2024deervladynamicinferencemultimodal, del2023skipdecode}, as shown in \cref{fig:teaser} (B), where discarding deeper layers risks losing critical semantic information. Inspired by the Shallow Brain Hypothesis (SBH)~\cite{suzuki2023deep}, which suggests that the brain balances deep hierarchical structures with shallow, parallel cortico-subcortical loops for cognition and causal reasoning, we propose a selective layer activation strategy in VLA models. As shown in ~\cref{fig:teaser} (C), our approach mirrors the brain's dynamic depth-parallelism balance, activating only task-relevant layers to enhance efficiency and adaptability, embodying principles of SBH in VLA model design.

In this paper, we introduce a \textbf{\underline{M}}ixture-\textbf{\underline{o}}f-\textbf{\underline{L}}ay\textbf{\underline{e}}rs \textbf{\underline{V}}ision-\textbf{\underline{L}}anguage-\textbf{\underline{A}}ction model (\textbf{\textit{MoLe-VLA}}) incorporating a novel layer-selection router at the input stage of LLMs for its sparsity. Our design emulates the brain's decision-making process described in the SBH by dynamically selecting optimal forward pathways with varying layer combinations. Inspired by the routing mechanism in mixture-of-experts (MoE)~\cite{lin2024moellavamixtureexpertslarge, zhang2024decomposing, zhang2024efficient}, which enables horizontal expert-wise activation within a single LLM layer, we extend this concept vertically to achieve layer-wise activation. Specifically, we treat each LLM layer as an independent expert and utilize a biologically inspired router to manage layer skipping, mimicking the brain's selective activation of cortico-subcortical loops. Unlike Mixture-of-Depth (MoD)~\cite{raposo2024mixtureofdepthsdynamicallyallocatingcompute}, which assigns input tokens to different experts and risks token-wise inconsistencies due to varying perception levels across layers, our proposed MoLe dynamically selects the most relevant layers while processing input features holistically.

Traditional MoE or MoD routers, which rely on simple linear layers, often fail to capture critical spatial-temporal information necessary for reasoning in dynamic, embodied intelligence tasks. To address this limitation, we propose the \textit{Spatial-Temporal Aware Router (STAR)}, which independently processes spatial features from visual inputs and temporal dependencies from textual inputs. By combining these essential properties into a unified representation, STAR aligns the selection of LLM layers with the demands of the current environment. STAR dynamically activates the most relevant layers by generating softmax probabilities for each layer and selecting the top-$k$ layers with the highest probabilities. By fully leveraging spatial-temporal information, STAR ensures accurate and efficient adaptation to the dynamic nature of embodied intelligence tasks, achieving optimal performance with reduced computational overhead.

Nonetheless, skipping certain layers inevitably reduces the cognitive expressiveness of the model. To address this, we propose \textit{Cognitive self-Knowledge Distillation (CogKD)}, a novel approach to preserve grasping ability while mitigating cognitive collapse. In CogKD, the original full-layer model serves as the teacher, while the MoLe layer-skipping model acts as the student. Inspired by~\cite{li2024cogactfoundationalvisionlanguageactionmodel}, we introduce a learnable \textit{cognition token}, which efficiently integrates visual tokens and language guidance to enhance comprehension of task demands and produce task-relevant action sequences. By analyzing the similarity between cognition tokens and student tokens, we identify tokens of interest (ToIs) that represent task-critical information the student needs to learn. These ToIs provide precise guidance for adaptively re-weighting the distillation process, ensuring the student model focuses on key cognitive features while consistently benefiting from the layer-skipping efficiency.


The effectiveness of MoLe in both performance and efficiency enhancement is demonstrated in real-world and RLBench simulation environments based on various VLA models against state-of-the-art baselines. Extensive robotic experiments show that MoLe reduces the computational costs by $\times 5.6$ while improving model performance by up to 8\%. The key contributions of this work are summarized as:
\begin{itemize}
    \item We draw inspiration from the Shallow Brain Hypothesis to develop a MoLe framework, which mimics the signal flow in the human brain and enables dynamic layer activation via a router to improve model efficiency.
    \item We propose a novel layer-decision router, STAR, which fully leverages the spatial-temporal information from robotic inputs to make more accurate activation decisions.
    \item We introduce a self-knowledge distillation paradigm, CogKD, to recover cognitive information lost due to layer-skipping in sparse LLMs, enhancing overall performance.
\end{itemize}

\section{Related works}
\label{sec:related}

\subsection{Vision language action model}
The remarkable success of LLMs~\citep{openai2024gpt4technicalreport, touvron2023llamaopenefficientfoundation, raffel2023exploringlimitstransferlearning, liu2019robertarobustlyoptimizedbert} and VLMs~\citep{karamcheti2024prismaticvlmsinvestigatingdesign, radford2021learningtransferablevisualmodels, liu2023visualinstructiontuning, alayrac2022flamingovisuallanguagemodel, li2022blipbootstrappinglanguageimagepretraining} has driven the rapid development of VLA models~\citep{kim2024openvlaopensourcevisionlanguageactionmodel, li2024cogactfoundationalvisionlanguageactionmodel, brohan2023rt2visionlanguageactionmodelstransfer}, which extend VLMs by incorporating action generation. VLA models aim to bridge the gap between perception and action, enabling machines to not only interpret and understand visual and textual inputs but also generate and execute actions based on that understanding~\citep{li2024visionlanguagefoundationmodelseffective, awadalla2023openflamingoopensourceframeworktraining}. By integrating visual and linguistic information, these models produce more complex, context-aware outputs tailored to real-world environments, advancing their applicability in dynamic and embodied intelligence tasks.
\subsection{Efficient multimodal large language models}
With the advancement of VLA models, improving inference efficiency has become a critical area of research. Existing efforts can be categorized into three main strategies: efficient architectural design, model compression, and dynamic networks. Liu et al.\cite{liu2024robomambaefficientvisionlanguageactionmodel} leverage the Mamba model\cite{gu2024mambalineartimesequencemodeling} to enable efficient fine-tuning and inference, achieving pose prediction speeds 7× faster than existing robotic MLLMs in both simulation and real-world experiments. Wang et al.\cite{wang2022efficientvlmfastaccuratevisionlanguage} utilizes a lightweight model with only 93M parameters while retaining 98.4\% of its performance and delivering a 2.2× speedup. Yue et al.\cite{yue2024deervladynamicinferencemultimodal} propose a dynamic inference framework with multi-exit architectures, allowing early computation termination based on task-specific requirements. However, existing early-exit methods often overlook the significance of the final layers, which carry greater semantic relevance to downstream tasks. Building on dynamic networks, our work integrates knowledge distillation to achieve a layer-skipping mechanism, optimizing model performance while reducing redundant computations.

\subsection{Sparse mixture-of-experts}
While activation sparsity has been widely explored~\cite{li2024emergence, zhang2024multi}, sparse MoE model architecture has shown significant advantages in LLMs. ~\cite{shazeer2017outrageouslylargeneuralnetworks} demonstrated their ability to efficiently utilize vast numbers of parameters by activating only a small portion of the computation graph during inference. In the LLMs and VLMs era, MoE has become a widely adopted and effective architecture~\cite{dai2024deepseekmoeultimateexpertspecialization, zhang2024decomposing, zhang2024efficient}. For example, ~\cite{lin2024moellavamixtureexpertslarge} achieves performance comparable to LLaVA-1.5-7B on various visual understanding benchmarks and even surpasses LLaVA-1.5-13B on the object hallucination benchmark, using only ~3B sparsely activated parameters. Additionally, ~\cite{raposo2024mixtureofdepthsdynamicallyallocatingcompute} employs a router to dynamically choose between computational paths, such as a standard block's computation or a residual connection. While our model shares similarities with ~\cite{raposo2024mixtureofdepthsdynamicallyallocatingcompute}, we differ by employing a router to select all standard block computations, enabling a more comprehensive approach to layer activation.
\begin{figure*}[t]
\centering
\includegraphics[width=0.95\linewidth]{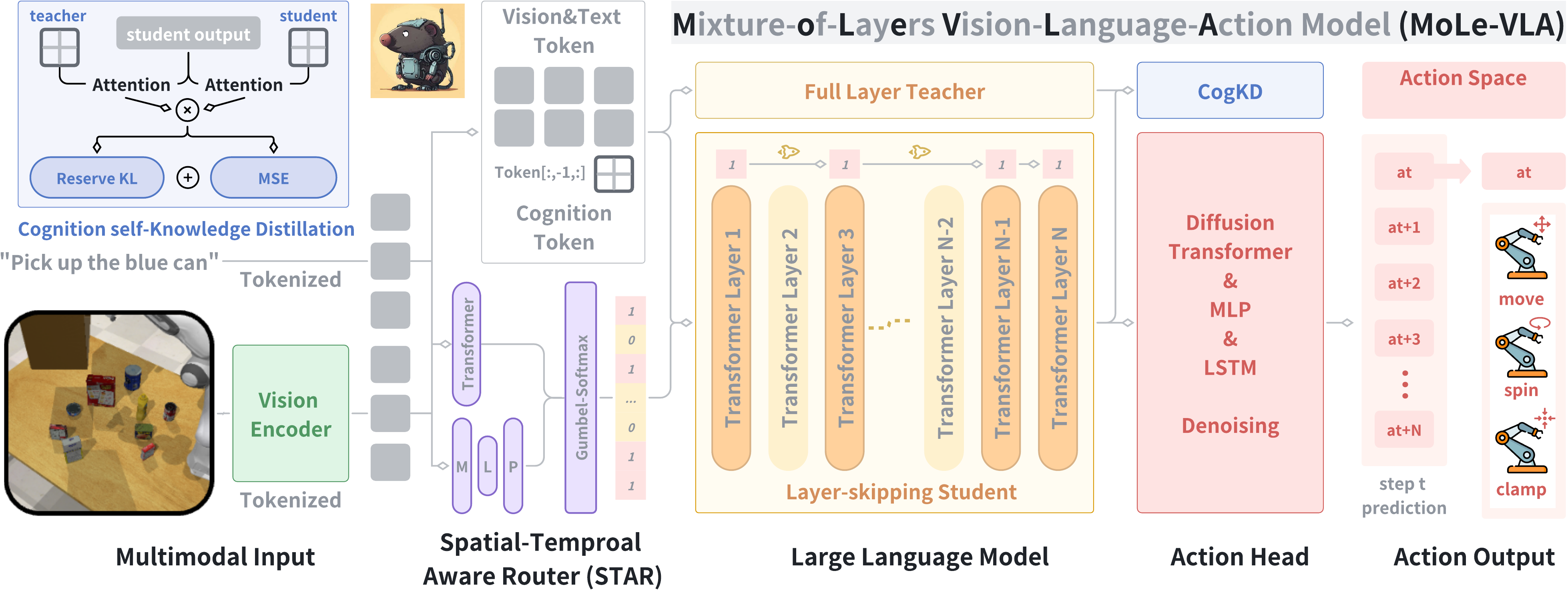}
\caption{\textbf{The overall framework of MoLe-VLA.} Our proposed \textit{Mixture of Layers (MoLe)} architecture consists of a \textit{Spatial-Temporal Aware Router (STAR)} and a devised \textit{Cognition self-Knowledge Distillation (CogKD)} for vision language action models.}
\label{fig:framework}
\end{figure*}

\definecolor{lightgray}{gray}{.9}
\definecolor{lightblue}{RGB}{230,240,255}
\definecolor{lightgreen}{RGB}{230,255,230}
\definecolor{lightyellow}{RGB}{255,255,230}
\definecolor{lightred}{RGB}{255,230,230}
\newcommand{\lb}[1]{\bf{\color{lightblue}{#1}}}
\newcommand{\ly}[1]{\bf{\color{lightyellow}{#1}}}
\newcommand{\lr}[1]{\bf{\color{lightred}{#1}}}

\definecolor{lightlightgray}{gray}{.95}
\definecolor{lightlightblue}{RGB}{240,245,255}
\definecolor{lightlightgreen}{RGB}{240,255,240}
\definecolor{lightlightyellow}{RGB}{255,255,240}
\definecolor{lightlightred}{RGB}{255,240,240}
\newcommand{\llb}[1]{\bf{\color{lightlightblue}{#1}}}
\newcommand{\llg}[1]{\bf{\color{lightlightgreen}{#1}}}
\newcommand{\lly}[1]{\bf{\color{lightlightyellow}{#1}}}
\newcommand{\llr}[1]{\bf{\color{lightlightred}{#1}}}

\definecolor{lightlightlightgray}{gray}{.99}
\definecolor{lightlightlightblue}{RGB}{247,250,255}
\definecolor{lightlightlightgreen}{RGB}{247,255,247}
\definecolor{lightlightlightyellow}{RGB}{255,255,247}
\definecolor{lightlightlightred}{RGB}{255,247,247}
\newcommand{\lllb}[1]{\bf{\color{lightlightlightblue}{#1}}}
\newcommand{\lllg}[1]{\bf{\color{lightlightlightgreen}{#1}}}
\newcommand{\llly}[1]{\bf{\color{lightlightlightyellow}{#1}}}
\newcommand{\lllr}[1]{\bf{\color{lightlightlightred}{#1}}}

\section{Methods}
\label{sec:methods}
\subsection{Preliminary: Mixture-of-Experts}
The MoE paradigm enhances model capacity while maintaining computational efficiency via conditional computation. For an input $\bm{x} \in \mathbb{R}^d$, a standard MoE layer is defined as:
\begin{equation}
\text{MoE}(\bm{x}) = \sum_{i=1}^{N_e} G_i(\bm{x}) \cdot \mathit{E}(\bm{x}),
\end{equation}
where $N_e$ is the number of experts, $\mathit{E}: \mathbb{R}^d \rightarrow \mathbb{R}^d$ represents the $i$-th expert network, and $G(\bm{x}) = \{G_1(\bm{x}), \dots, G_{N_e}(\bm{x})\}$ is the gating function satisfying $\sum_{i=1}^{N_e} G_i(\bm{x}) = 1$. The gating weights are computed as:
\begin{equation}
G(\bm{x}) = \text{Softmax}(\mathbf{W}_g \cdot \bm{x} + \mathbf{b}_g),
\end{equation}
where $\mathbf{W}_g \in \mathbb{R}^{N_e \times d}$ and $\mathbf{b}_g \in \mathbb{R}^{N_e}$ are learnable parameters. To improve efficiency, sparse gating with top-$k$ selection is often applied. To address load imbalance, where too many inputs are routed to a few experts, a load balance loss $\mathcal{L}_{\text{lb}}$ is introduced:
\begin{equation}
\mathcal{L}_{\text{lb}} = \frac{1}{N_e} \sum_{i=1}^{N_e} \bigg( \frac{\sum_{n=1}^N v_i(\bm{x}_n)}{\sum_{n=1}^N v_i(\bm{x}_n) + \epsilon} \bigg)^2,
\end{equation}
where $v_i(\bm{x}_n) = 1$ if the $i$-th expert is selected for input $\bm{x}_n$ by the top-$k$ gating mechanism, and $v_i(\bm{x}_n) = 0$ otherwise. This loss encourages balanced expert utilization and improves computational efficiency.

\subsection{Mixture-of-Layers: MoLe-VLA}
\textbf{Vision language action model.} Tasked with a language instruction $\bm{l}$ with a length $L$, a robot receives an observation $o_t$ from sensors (\emph{e.g.}, RGB image from the camera) at timestep $t$ to predict the action space of a gripper with 7 degrees of freedom (DoF) to execute:
\begin{equation}
    \mathbf{a}_t^*=[\Delta x,\Delta y, \Delta z, \Delta\phi, \Delta\theta, \Delta\psi, g],
\end{equation} 
where $\Delta x,\Delta y, $ and $\Delta z$ are the relative translation offsets of the end effector, $\Delta\phi, \Delta\theta$, $\Delta\psi$ denote the rotation changes, and $g\in\{0,1\}$ indicates the gripper open/close state.

Our basic VLA model mainly consists of a vision encoder $\mathcal{E}$, an MLLM $\pi$, and an action module $\mathcal{A}$.
The vision encoder $\mathcal{E}$ comprises DINO-v2~\cite{oquab2023dinov2} and Siglip~\cite{zhai2023sigmoid}, which encodes an input image $o_t$ into a sequence of informative tokens $v_t$. For multimodal fusion, an MLLM is established on top of the visual representations generated by the vision encoder $\mathcal{E}$, which functions as an effective multimodal feature extractor $\pi$, formalized as follows:
\begin{equation}
\label{eqn:MLLM}
\bm{f}_t =\pi(\bm{l}, \mathcal{E}(o_t)),
\end{equation}
where the output $\bm{f}_t$ represents the hidden state sequence from the last layer of our MLLM at timestep $t$, corresponding to the cognition token. This serves as a condition for the subsequent action module to interpret and derive the desired actions. Folllowing CogAct~\cite{li2024cogactfoundationalvisionlanguageactionmodel}, our action module $\mathcal{A}$ takes the cognition feature $\bm{e}_t^{c}$ extracted from the output feature $\bm{f}_t$ as input and predicts the final actions $\mathbf{a}_t^*$.

Our vision, language, and action modules are trained end-to-end by minimizing the mean squared error between the predicted noises from the action module and the ground truth noises. Taking the diffusion head as an example, the loss function is defined as:
\begin{equation}
    \mathcal{L}_{task} = \mathbb{E}_{\epsilon\sim\mathcal{N}(0,1),i}||\hat{\epsilon}^{i}-\epsilon||
\end{equation}
where $\hat{\epsilon}^{i}$ is the predicted noise for the noisy action $a_{t}^*$ at the $i$'s denoising step, and $\epsilon$ is the corresponding ground truth.

\textbf{Layer-skipping mechanism via MoLe router.}
We propose MoLe-VLA to improve the efficiency of LLM in robotic tasks, where many transformer layers are underutilized due to the simpler reasoning demands of robotics tasks. MoLe employs a lightweight router to adaptively skip non-essential transformer layers during inference, reducing computational costs while maintaining performance.

As shown in ~\cref{fig:framework}, for a given MLLM $\pi$ with $K$ layers, the MoLe router processes the input embeddings $\bm{x}_{k} \in \mathbb{R}^{b \times n \times d}$ and generates a binary gating vector $G_{mol}(\bm{x}) = \{G_{k}\}_{k=1}^{K}$, where $G_k \in [0, 1]$. To ensure efficiency, only the top-$k$ values in $G_{mol}(\bm{x})$ are set to 1, determining which layers $\pi_{k}$ are executed with the hidden feature $\bm{h}_{k}$ while the rest are skipped:
\begin{equation}
    \bm{h}_{k} = G_{k} \cdot \pi_{k}(\bm{h}_{k-1}) + (1-G_{k}) \cdot \bm{h}_{k-1}.
\end{equation}

Unlike traditional MoE routers that allocate tokens to experts, the MoLe router skips entire layers, avoiding redundant computations. This improves inference efficiency and responsiveness, making MoLe particularly suited for real-time robotic tasks like manipulation and navigation that require lightweight and adaptive processing. The complete pseudo-code of MoLe is provided in Algorithm \ref{alg:mole}.

\subsection{Spatial-Temporal Aware Router}
We propose a novel routing mechanism that synergistically leverages the spatial structure of visual inputs and the temporal dependencies in language inputs to select appropriate LLM layers for VLA tasks dynamically.  
Given visual features $\bm{v}_{t} \in \mathbb{R}^{b \times n_{img} \times d}$ and textual features $\bm{l} \in \mathbb{R}^{b \times n_{text} \times d}$, both modalities are projected into a shared latent space using a learnable matrix $\mathbf{W}_p \in \mathbb{R}^{d \times d_1}$:  
\begin{equation}
    \bm{h}_{img} = \bm{v}_{t} \cdot \mathbf{W}_p, \quad \bm{h}_{text} = \bm{l} \cdot \mathbf{W}_p.
\end{equation}

We compute spatial routing weights $\mathbf{S} \in \mathbb{R}^{b \times N_e}$ from $\bm{h}_{img}$ to capture spatial features:  
\begin{equation}
    \mathbf{S} = \mathbf{W}_s^{(2)} \cdot \varphi(\mathbf{W}_s^{(1)} \cdot \bm{h}_{img} + \mathbf{b}_s^{(1)}), 
\end{equation}
where $\varphi$ is the GELU activation. Concurrently, temporal routing weights $\mathbf{T} \in \mathbb{R}^{b \times N_e}$ are derived from $\bm{h}_{text}$ using a Transformer module, followed by average pooling:  
\begin{equation}
    \mathbf{T} = \mathbf{W}_t \cdot \Phi(\text{Transformer}(\bm{h}_{text})).
\end{equation}

A dynamic temperature factor $\alpha \in [0,1]$, computed from the [CLS] token of $\bm{h}_{text}$, modulates routing sharpness:  
\begin{equation}
    \alpha = \sigma(\mathbf{W}_\tau^\top \cdot \bm{h}_{text}^{[CLS]} + b_\tau),
\end{equation}
where $\sigma$ is the sigmoid function. The final expert gating weights $\mathbf{G} \in \mathbb{R}^{b \times N_e}$ combine $\mathbf{S}$ and $\mathbf{T}$, scaled by $\alpha$, and are computed via Gumbel-Softmax for differentiable selection:  
\begin{equation}
    \mathbf{G} = \tau(\alpha \cdot (\mathbf{S} + \mathbf{T}), \tau=1.0).
\end{equation}

By integrating spatial and temporal information, our method enables the router to select LLM layers, optimizing performance for VLA tasks adaptively. The approach is efficient, requiring only $\mathcal{O}(N_e(d_2 + N_{text}^2))$ FLOPs per sample compared to $\mathcal{O}(N_e d)$ in standard MoE frameworks, where $d \gg N_{text}, d_2$. This design ensures high adaptability and computational efficiency.

\begin{algorithm}[t]
    \SetAlgoLined
    \KwInput{\textcolor{black}{Observation $o_{t}$, language instruction $\bm{l}$, ground truth $\bm{\epsilon}$, total layers $K$}}
    \KwOutput{\textcolor{black}{Student model loss $\mathcal{L}_{MoLe}$}}
    Obtain visual tokens: $\bm{v}\gets\mathcal{E}(o_t)$;
    
    Concat multimodal token: $\bm{x}\gets concat(\bm{v}, \bm{l})$;
    
    \vspace{0.5em}
    \textbf{\protect\textcolor{teal}{Step 1: Compute skip indices via STAR router}}
    Compute skip indices: $\{g_{k}\}_{k=1}^{K},\ \mathcal{L}_{lb} \gets G_{star}(\bm{v}, \bm{l})$\;
    
    \vspace{0.5em}
    \textbf{\protect\textcolor{teal}{Step 2: Compute skip indices via STAR router}}
    \For{$k = 1$ \KwTo $K$ in $\pi^{(s)}(\bm{x})$}{
        $\bm{h}_{k} \gets 
        \begin{cases} 
        \pi_{k}^{(s)}(\bm{h}^{(s)}_{k-1}) & \text{\textcolor{brown}{if }} G_k = 1 \\[0.5em]
        \bm{h}^{(s)}_{k-1} & \text{\textcolor{brown}{otherwise}}
        \end{cases}$\;
    }
    Final student features: $\bm{f}^{(s)} \gets \bm{h}^{(s)}_K$ \;
    Cognition tokens: $\bm{e}^{c,(s)} \gets \bm{h}^{(s)}_K[:, -1, :]$\;
    
    \vspace{0.5em}
    \textbf{\protect\textcolor{teal}{Step 3: Compute skip indices via STAR router}}
    Execute all layers in $\pi^{(t)}(\bm{x})$: $\bm{f}^{(t)} \gets \bm{h}_K^{(t)}$\;
    Compute CogKD loss: $\mathcal{L}_{cog}\gets \mathcal{L}_{mse} + \mathcal{L}_{reservekl}$;
    
    \vspace{0.5em}
    \protect\textbf{\protect\textcolor{teal}{Step 4: Compute skip indices via STAR router}}
    Compute action prediction: $\hat{\epsilon} \gets \mathcal{A}(\bm{f}^{(s)})$\;
    Compute loss: $\mathcal{L}_{MoLe} \gets \mathcal{L}_{task} + \mathcal{L}_{cog} + \mathcal{L}_{lb}$
    
    \vspace{0.5em}
    \KwRet $\mathcal{L}_{MoLe}$\;
    
\caption{\textsc{\textcolor{black}{MoLe with STAR and CogKD}}}
\label{alg:mole}
\end{algorithm}

\subsection{Cognition self-Knowledge Distillation}
\begin{figure}[t]
\includegraphics[width=0.48\textwidth]{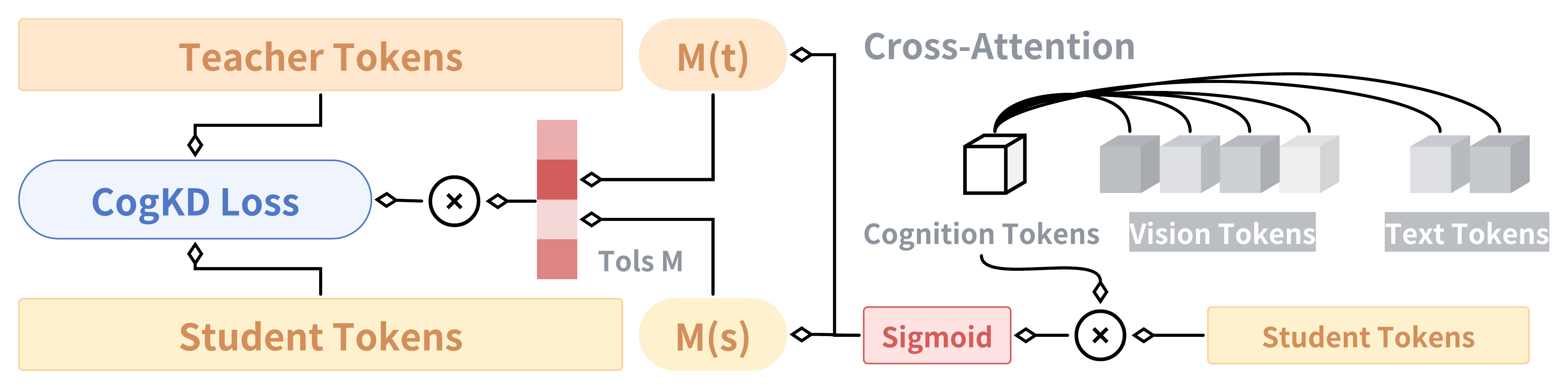}
\centering
\caption{\textbf{Detailed illustration of our proposed CogKD loss.}}
\label{fig:kd}
\end{figure}

While achieving an efficient layer-skipping mechanism, we also design a self-distillation strategy to compensate for the cognition loss in the sparse LLM as shown in \cref{fig:kd}. Here, we take the original model as the teacher and the MoLe model as the student. To distill the tokens, one common approach is to mimic the tensor token-wisely \citep{yang2024vitkd, zhang2024freekd, cao2025move}. Formally, with the tokens $\bm{f}^{(t)}\in\mathbb{R}^{n\times d}$ and $\bm{f}^{(s)}\in\mathbb{R}^{n\times d_s}$ of teacher and student networks, the mimicking can be fulfilled via token reconstruction as 
\begin{equation} \label{eq:mimic}
    \mathcal{L}_\mathrm{mimic} = \frac{1}{N} \left\|\bm{f}^{(t)} - \mu(\bm{f}^{(s)})\right\|_2^2 ,
\end{equation}
However, the Eq.\ref{eq:mimic} \textit{treats and distills each token equally}, which is inappropriate. For instance, the visual tokens related to the text description should receive more attention \cite{zhang2024sparsevlm, ye2024fit}. 

Therefore, we introduce a learnable embedding $\bm{e}_{t}^{c}\in\mathbb{R}^{1\times d}$ dubbed \textit{cognition token} to distill adaptively. Specifically, it is inserted in the bottom layer, effectively integrating vision tokens and language instruction to understand task requirements better and generate action sequences relevant to the task. The teacher and student model each have their own $\bm{e}_{t}^{c,(t)}$ and $\bm{e}_{t}^{c,(s)}$, respectively. During the distillation, we get the tokens of interests (ToIs) $\bm{M}$ by calculating the similarity between the cognition token and the student tokens:
\begin{equation}
    \bm{M}^{(i)} = \eta(\bm{e}^{c,(i)}\bm{f}^{(s)}), i \in \{s,t\},
\end{equation}
where $\eta$ denotes the Sigmoid function. Next, we utilize the \textit{intersection} of ToIs generated by the teacher and student cognition tokens to decide the distillation degree of each token, where $\bm{M} = \bm{M}^{(t)} \odot \bm{M}^{(s)}$, because the distillation tokens should consist of the ones both important to the teacher and student. Therefore, the Eq.\ref{eq:mimic} can be updated as:
\begin{equation} \label{eq:cog_mimic}
    \mathcal{L}_\mathrm{cog\text{-}mimic} = \frac{1}{N} \left\|\bm{M} \odot \bm{f}^{(t)} - \mu(\bm{M} \odot \bm{f}^{(s)})\right\|_2^2.
\end{equation}
Furthermore, we introduce the \textit{Reverse-KL} \cite{gu2023minillm} paired with our cognition token as the before manner to obtain $\mathcal{L}_\mathrm{cog\text{-}reversekl}$ to enhance distribution constraint:
\begin{equation} \label{eq:cog_rkl}
    \mathcal{L}_\mathrm{cog\text{-}reversekl} = (\bm{M} \odot \bm{f}^{(s)}) \log\left(\frac{\bm{M} \odot \bm{f}^{(s)}}{\bm{M} \odot \bm{f}^{(t)}}\right).
\end{equation}
Finally, our eventual CogKD loss can be formulated as
\begin{equation} \label{eq:overall_loss}
    \mathcal{L}_\mathrm{cog} = (1-\lambda_1)\mathcal{L}_\mathrm{cog\text{-}mimic} + \lambda_1\mathcal{L}_\mathrm{cog\text{-}reversekl},
\end{equation}
where $\lambda_1$ is the factor and set to $0.5$ for balancing the losses.

\newcolumntype{C}[1]{>{\columncolor{#1}}c}
\begin{table*}[tb]
\caption{\textbf{Performance comparison with existing VLA models across ten tasks in RLBench settings.} We colour-coded the results \textbf{\textbf{red}} \textbf{(1st)} and \textcolor{blue}{blue} (2nd) and the row colour reflects the baseline type. The five efficiency methods operate with only 50\% LLM layers.} 
\label{tab:rlbench}
\resizebox{\linewidth}{!}{%
\setlength{\tabcolsep}{2mm}{
\begin{tabular}{cccccccc}
\toprule\toprule
\multirow{2}{*}{\textbf{Methods}} & \multirow{2}{*}{Action Head} & \multirow{2}{*}{Backbone} & \cellcolor{lightgray}Put Rubbish  & \cellcolor{lightgray}Close & \cellcolor{lightgray}Close & \cellcolor{lightgray}Take Umbrella & \cellcolor{lightgray}Close \\
 & & & \cellcolor{lightgray}in Bin & \cellcolor{lightgray}Box & \cellcolor{lightgray}Laptop Lid & \cellcolor{lightgray}out of Stand & \cellcolor{lightgray}Fridge\\
\midrule\midrule
\rowcolor{lightblue}\multicolumn{8}{c}{\textbf{\textit{RLBench}}} \\ 
\midrule
\rowcolor{lightlightlightgreen}
\multicolumn{1}{c}{\cellcolor{lightgreen}OpenVLA~\cite{kim2024openvlaopensourcevisionlanguageactionmodel}~\textcolor{gray}{(CoRL'24)}} & MLP & LLaMA2-7B & 8.0\% & 72.0\% & 64.0\% & \textit{28.0\%} & 88.0\% \\
\rowcolor{lightlightlightgreen}
\multicolumn{1}{c}{\cellcolor{lightgreen}CogAct~\cite{li2024cogactfoundationalvisionlanguageactionmodel}~\textcolor{gray}{(Arxiv'24)}} & Diffusion & LLaMA2-7B & \textbf{60.0\%} & 64.0\% & \textit{76.0\%} & 32.0\% & 48.0\% \\
\rowcolor{lightlightlightyellow}
\multicolumn{1}{c}{\cellcolor{lightyellow}RoboMamba~\cite{liu2024robomambaefficientvisionlanguageactionmodel}~\textcolor{gray}{(NeruIPS'24)}} & MLP & Mamba-2.8B & \textit{36.0\%} & 60.0\% & 52.0\% & 32.0\% & 68.0\% \\
\rowcolor{lightlightlightyellow}
\multicolumn{1}{c}{\cellcolor{lightyellow}Random-skip-CogAct} & MLP & LLaMA2-7B & 16.0\% & \textit{80.0\%} & \textbf{80.0\%} & 32.0\% & 84.0\% \\
\rowcolor{lightlightlightyellow}
\multicolumn{1}{c}{\cellcolor{lightyellow}MoD-CogAct~\cite{raposo2024mixtureofdepthsdynamicallyallocatingcompute}~\textcolor{gray}{(Arxiv'24)}} & Diffusion & LLaMA2-7B & \textit{56.0\%} & \textit{80.0\%} & 68.0\% & \textbf{40.0\%} & \textit{92.0\%} \\
\rowcolor{lightlightlightyellow}
\multicolumn{1}{c}{\cellcolor{lightyellow}DeeR-CogAct~\cite{yue2024deervladynamicinferencemultimodal}~\textcolor{gray}{(NeruIPS'24)}}  & Diffusion & LLaMA2-7B & 52.0\% & 72.0\% & 60.0\% & \textit{36.0\%} & 76.0\% \\
\rowcolor{lightlightlightred}
\multicolumn{1}{c}{\cellcolor{lightred}MoLe-OpenVLA~\textcolor{gray}{(Ours)}}  & MLP & LLaMA2-7B & 12.0\% & \textit{80.0\%} & \textit{76.0\%} & \textbf{40.0\%} & \textbf{96.0\%} \\
\rowcolor{lightlightlightred}
\multicolumn{1}{c}{\cellcolor{lightred}MoLe-CogAct~\textcolor{gray}{(Ours)}}  & Diffusion & LLaMA2-7B & 24.0\% & \textbf{84.0\%} & \textbf{80.0\%} & \textit{36.0\%} & 88.0\% \\
\midrule
\multirow{2}{*}{\textbf{Methods}} & \cellcolor{lightgray}Sweep & \cellcolor{lightgray}Phone & \cellcolor{lightgray}Change  & \cellcolor{lightgray}Toilet & \cellcolor{lightgray}Take Frame & \multirow{2}{*}{Mean Acc.\% $\uparrow$} &  \multirow{2}{*}{FLOPs (G) $\downarrow$} \\
& \cellcolor{lightgray}to Dustpan & \cellcolor{lightgray}on Base & \cellcolor{lightgray}Clock & \cellcolor{lightgray}Seat Down & \cellcolor{lightgray}off Hanger &  & \\
\midrule

\rowcolor{lightlightlightgreen}
\multicolumn{1}{c}{\cellcolor{lightgreen}OpenVLA~\cite{kim2024openvlaopensourcevisionlanguageactionmodel}~\textcolor{gray}{(CoRL'24)}} & \textit{68.0\%} & 20.0\% & \textit{16.0\%} & 76.0\% & 12.0\% & 45.4\% & 1930.0 \\
\rowcolor{lightlightlightgreen}
\multicolumn{1}{c}{\cellcolor{lightgreen}CogAct~\cite{li2024cogactfoundationalvisionlanguageactionmodel}~\textcolor{gray}{(Arxiv'24)}} & 44.0\% & \textit{56.0\%} & 12.0\% & \textbf{100.0\%} & 60.0\% & 57.2\% & 1935.8 \\
\rowcolor{lightlightlightyellow}
\multicolumn{1}{c}{\cellcolor{lightyellow}RoboMamba~\cite{liu2024robomambaefficientvisionlanguageactionmodel}~\textcolor{gray}{(NeruIPS'24)}} & 32.0\% & 44.0\% & \textit{16.0\%} & 64.0\% & 32.0\% & 43.6\% & \textcolor{red}{826.3} \\
\rowcolor{lightlightlightyellow}
\multicolumn{1}{c}{\cellcolor{lightyellow}Random-skip-CogAct} & 64.0\% & 24.0\% & 8.0\% & 92.0\% & 32.0\% & 51.2\%\small\textcolor{gray}{(-6.0\%)} & 984.3 \\
\rowcolor{lightlightlightyellow}
\multicolumn{1}{c}{\cellcolor{lightyellow}MoD-CogAct~\cite{raposo2024mixtureofdepthsdynamicallyallocatingcompute}~\textcolor{gray}{(Arxiv'24)}} & 4.0\% & 36.0\% & \textbf{20.0\%} & \textit{96.0\%} & \textbf{72.0\%} & 56.4\%\small\textcolor{gray}{(-0.8\%)} & 985.8 \\
\rowcolor{lightlightlightyellow}
\multicolumn{1}{c}{\cellcolor{lightyellow}DeeR-CogAct~\cite{yue2024deervladynamicinferencemultimodal}~\textcolor{gray}{(NeruIPS'24)}}  & 36.0\% & \textbf{68.0\%} & \textbf{20.0\%} & \textit{96.0\%} & \textit{68.0\%} & \textcolor{blue}{\textit{59.2\%}}\small\textcolor{gray}{(+2.0\%)} & 997.4 \\
\rowcolor{lightlightlightred}
\multicolumn{1}{c}{\cellcolor{lightred}MoLe-OpenVLA~\textcolor{gray}{(Ours)}}  & \textbf{72.0\%} & 20.0\% & 12.0\% & \textbf{100.0\%} & 44.0\% & 55.6\%\small\textcolor{gray}{(+10.2\%)} & \textcolor{blue}{981.5} \\
\rowcolor{lightlightlightred}
\multicolumn{1}{c}{\cellcolor{lightred}MoLe-CogAct~\textcolor{gray}{(Ours)}}  & \textit{68.0\%} & 36.0\% & \textbf{20.0\%} & \textbf{100.0\%} & \textbf{\textbf{72.0\%}} & \textcolor{red}{\textbf{60.8\%}}\small\textcolor{gray}{(+3.6\%)} & 985.8 \\
\bottomrule\bottomrule
\end{tabular}%
}}
\end{table*}

\subsection{Optimization Objective}
For the update of the teacher model, we initialize both models with pre-trained parameters and use the exponential moving average (EMA) to update the teacher model $\pi^{(t)}$:
\begin{equation}
\pi^{(t)}_{t} = \alpha \cdot\pi^{(t)}_{t-1} + (1-\alpha) \cdot\pi^{(s)}_{t}.
\label{eq:ema}
\end{equation}
In this setup, $t$ indicates the time step, and we set the update weight $\alpha = 0.999$ \cite{AnttiTarvainenetal2017}.

Our final training objective can be formulated with the combination of $\mathcal{L}_{task}$, $\mathcal{L}_{cog}$ and $\mathcal{L}_{lb}$:
\begin{equation}
    \mathcal{L}_{MoLe} = \mathcal{L}_{task} + \lambda_{2}\mathcal{L}_{cog} + \lambda_{3}\mathcal{L}_{lb},
\end{equation}
where $\lambda_2$ and $\lambda_3$ are two hyperparameters which are set to 0.5 and 0.1 by default. A more detailed discussion about the hyperparameters can be found in the Appendix.
\section{Experiments}
\label{sec:experiments}
\definecolor{lightgray}{gray}{.9}
\definecolor{lightblue}{RGB}{230,240,255}
\definecolor{lightgreen}{RGB}{230,255,230}
\definecolor{lightyellow}{RGB}{255,255,230}
\definecolor{lightred}{RGB}{255,230,230}

\definecolor{lightlightgray}{gray}{.95}
\definecolor{lightlightblue}{RGB}{240,245,255}
\definecolor{lightlightgreen}{RGB}{240,255,240}
\definecolor{lightlightyellow}{RGB}{255,255,240}
\definecolor{lightlightred}{RGB}{255,240,240}

\definecolor{lightlightlightgray}{gray}{.99}
\definecolor{lightlightlightblue}{RGB}{247,250,255}
\definecolor{lightlightlightgreen}{RGB}{247,255,247}
\definecolor{lightlightlightyellow}{RGB}{255,255,247}
\definecolor{lightlightlightred}{RGB}{255,247,247}


\subsection{Implementation details}
\label{sec:details}
\paragraph{Simulation and real-world deployment.} To evaluate our approach and demonstrate its generalization ability, we conduct experiments on both RLBench~\cite{james2020rlbench} in the CoppeliaSim simulator and real-world environments with :

\textit{1) RLBench} includes 10 diverse tabletop tasks performed with a Franka Panda robot and a front-view camera. These tasks range from object manipulation to environment interaction, such as: \textit{Close box}, \textit{Close laptop lid}, \textit{Toilet seat down}, \textit{Put rubbish in bin}, \textit{Sweep to dustpan}, \textit{Close fridge}, \textit{Phone on base}, \textit{Take umbrella out of stand}, \textit{Frame off hanger}, and \textit{Change clock}. Task data are generated using predefined waypoints and the Open Motion Planning Library~\cite{sucan2012open}. Following prior work~\cite{jia2024lift3d}, each task includes 100 training trajectories sampled using a frame-based approach and evaluated in 25 trials per task within the training workspace.

\textit{2) Real-world deployment} is evaluated on the Franka Research 3 (FR3) robot equipped with a 3D-printed UMI gripper~\cite{chi2024universal} across three tasks. A GoPro 9 camera mounted on the wrist captures real-world visual observations. We collect 50 demonstrations for each task, including \textit{detach charger}, \textit{pull drawer}, and \textit{pour water}, using a hand-held UMI gripper within a defined workspace range. A single agent is trained across all tasks and evaluated in 10 trials per task within the training workspace. The success rate is determined through human assessment and serves as the evaluation metric.

\begin{figure*}[t]
\centering
\includegraphics[width=1\linewidth]{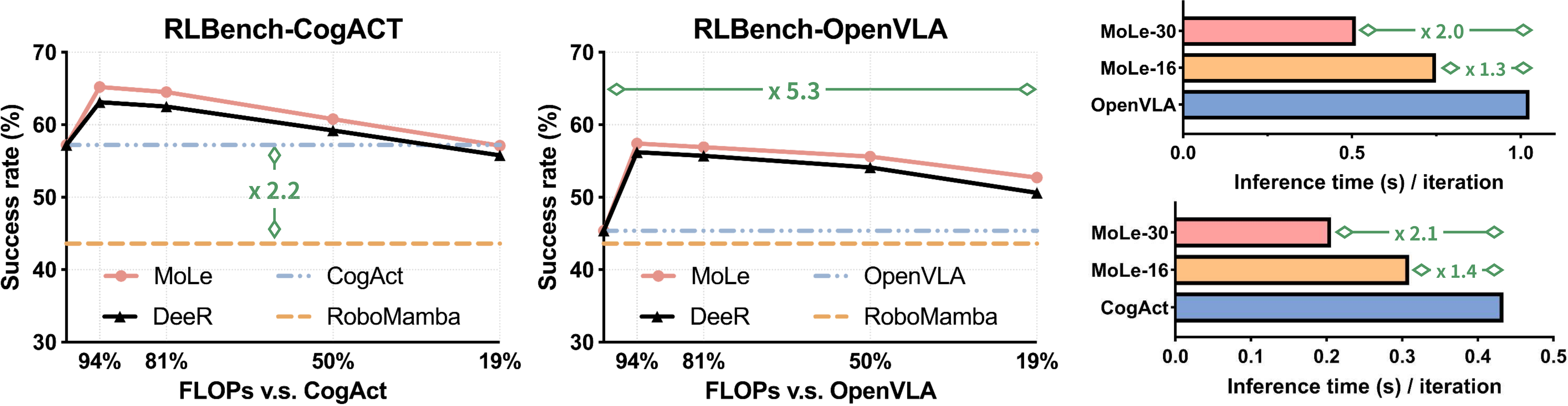}
\caption{\textbf{Efficiency analysis compared with state-of-the-art baselines with FLOPs and inference time.} (Left) Success rate \textit{v.s.} the FLOPs reduction compared to model backbone. (Right) Inference time per iteration for different layers of MoLe and model backbones.}
\label{fig:efficiency}
\vspace{-0.2cm}
\end{figure*}

\vspace{-2em}
\paragraph{Baselines} The innovation of \textit{MoLe-VLA} lies in its novel, plug-in MoLe architecture, which accelerates VLA inference while improving the robot’s success rate. To evaluate its effectiveness, we compare \textit{MoLe} with three state-of-the-art VLA methods across two action generation paradigms: 1) \textit{Autoregressive models}, including \textit{OpenVLA}~\cite{kim2024openvlaopensourcevisionlanguageactionmodel}, which uses LLaMA for discrete action prediction, and 2) \textit{Diffusion-based models}, such as \textit{CogAct}~\cite{li2024cogactfoundationalvisionlanguageactionmodel}, which predicts action chunks via a diffusion head. Additionally, we evaluate several VLA efficiency baselines: \textit{RoboMamba}~\cite{liu2024robomambaefficientvisionlanguageactionmodel}, which replaces transformer-based LLMs with a lightweight Mamba model; \textit{DeeR}~\cite{yue2024deervladynamicinferencemultimodal}, which enables early exits in LLMs; \textit{MoD}~\cite{raposo2024mixtureofdepthsdynamicallyallocatingcompute}, which allocates input tokens dynamically across layers; and \textit{Random-skip}, which skips LLM layers randomly. For a fair comparison, the latter three baselines are implemented on CogAct with the same setting, with \textit{DeeR} using single-phase training and full model loading. We integrate MoLe with two VLA models, forming \textit{MoLe-OpenVLA} and \textit{MoLe-CogAct}, both using a default \textbf{50\% layer-skip}.

\paragraph{Training and evaluation details.} All baselines are trained using the same task configuration for fair comparison. Each method's official pre-trained parameters are loaded, following their respective training settings. For MoLe-VLA, the single-view RGB input is resized to $224 \times 224$, and the robot state is aligned with the predicted actions (7-DOF end-effector poses). The model is trained with a batch size of 64 and 8 diffusion steps per sample, using pre-trained weights for the vision and language modules. The vision module incorporates \textit{DINO-v2} and \textit{SigLIP}, while the language module \textit{LLAMA-2} and the action module \textit{DiT-Base} are trained end-to-end with a constant learning rate of $2 \times 10^{-5}$ for 1k iterations. Training is conducted on 8 NVIDIA A800 GPUs in approximately 1.5 hours using PyTorch’s Fully Sharded Data Parallel (FSDP) framework.

\begin{table}[t]
\begin{center}
    \setlength\tabcolsep{10pt} 
    \caption{\label{tab:efficiency} \textbf{Inference analysis} evaluated on RLBench simulation environment with FLOPs, inference time, and mean success rate.}
        \vspace{-0.5em}
    \resizebox{0.99\columnwidth}{!}{%
        \begin{tabular}{cccc}
        \toprule\toprule
        \cellcolor{lightgray}\textbf{Methods} & \cellcolor{lightyellow}\textbf{Inference time$\downarrow$} & \cellcolor{lightgreen}\textbf{FLOPs (G)$\downarrow$} & \cellcolor{lightblue}\textbf{Mean$\uparrow$} \\
        \midrule\midrule
        
        
        \cellcolor{lightlightlightgray}CogAct   & \cellcolor{lightlightlightyellow}0.434 s  & \cellcolor{lightlightlightgreen}1935.8  & \cellcolor{lightlightlightblue}57.2\% \\ 

        \cellcolor{lightlightlightgray}DeeR    & \cellcolor{lightlightlightyellow}0.337 s & \cellcolor{lightlightlightgreen}997.4  & \cellcolor{lightlightlightblue}59.2\% \\ 
                
        \cellcolor{lightlightlightgray}MoLe   & \cellcolor{lightlightlightyellow}0.309 s & \cellcolor{lightlightlightgreen}985.8 & \cellcolor{lightlightlightblue}60.8\% \\ 

        \bottomrule\bottomrule
        \end{tabular}
    }
\end{center}
\vspace{-1.5em}
\end{table}

\begin{table}[t]
\begin{center}
    \setlength\tabcolsep{5pt} 
    \caption{\label{tab:quant} \textbf{Effective of model quantization} evaluated on RLBench simulation environment with NVIDIA 4090D GPU.}
    \vspace{-0.5em}
    \resizebox{0.99\columnwidth}{!}{%
        \begin{tabular}{ccccc}
        \toprule\toprule
        \cellcolor{lightgray}\textbf{Methods} & \cellcolor{lightred}\textbf{Precision} & \cellcolor{lightgreen}\textbf{Frequency$\uparrow$} & \cellcolor{lightyellow}\textbf{GPU memory$\downarrow$} & \cellcolor{lightblue}\textbf{Mean$\uparrow$} \\
        \midrule\midrule

        \cellcolor{lightlightlightgray}CogAct    & \cellcolor{lightlightlightred}FP16 & \cellcolor{lightlightlightgreen}9.8 Hz & \cellcolor{lightlightlightyellow}16055 MB & \cellcolor{lightlightlightblue}57.2\% \\ 

        \cellcolor{lightlightlightgray}MoLe    & \cellcolor{lightlightlightred}INT8 & \cellcolor{lightlightlightgreen}15.7 Hz & \cellcolor{lightlightlightyellow}8887 MB & \cellcolor{lightlightlightblue}58.8\%  \\ 

        \bottomrule\bottomrule
        \end{tabular}
    }
\end{center}
\vspace{-2em}
\end{table}

\subsection{Quantitative results in simulation.} 
\label{sec:simulation}
\paragraph{Performance enhancement}
We compare the performance of our proposed MoLe method with state-of-the-art VLA models across ten RLBench tasks, utilizing only half of the LLM layers for efficiency, as shown in \cref{tab:rlbench}. MoLe, implemented with OpenVLA and CogAct backbones, achieves superior success rates and efficiency. Notably, MoLe-CogAct achieves the highest mean success rate of \textit{60.8\%}, outperforming competing efficiency methods like DeeR of \textit{59.2\%} and MoD of \textit{56.4\%} as they overlook the most semantic layers and result in token-wise perception inconsistency, with significant improvements in tasks such as \textit{Close Fridge} and \textit{Sweep to Dustpan}. Similarly, MoLe-OpenVLA demonstrates a \textit{10.2\%} improvement over the original OpenVLA. Despite requiring only \textit{981.5} and \textit{985.8} GFLOPs, MoLe surpasses DeeR and MoD in efficiency and success rate, highlighting its ability to balance computational cost and task performance. These results underscore MoLe's effectiveness as a plug-in LLM architecture for robotic manipulation.

\begin{table}[t]
\begin{center}
    \setlength\tabcolsep{6pt} 
    \caption{\label{tab:scalability} \textbf{Scalability analysis} with mean success rate evaluated on RLBench simulation environment with different model sizes.}
        \vspace{-0.5em}
    \resizebox{0.99\columnwidth}{!}{%
        \begin{tabular}{cccc}
        \toprule\toprule
        \cellcolor{lightgray}\textbf{Methods} & \cellcolor{lightred}\textbf{CogAct-Small} & \cellcolor{lightyellow}\textbf{CogAct-Base} & \cellcolor{lightblue}\textbf{CogAct-Large} \\
        \midrule\midrule

        \cellcolor{lightlightlightgray}CogAct   & \cellcolor{lightlightlightred}47.2\% & \cellcolor{lightlightlightyellow}57.2\% & \cellcolor{lightlightlightblue}70.0\%  \\ 
                
        \cellcolor{lightlightlightgray}MoLe   & \cellcolor{lightlightlightred}49.9\%\textcolor{gray}{(+2.7\%)} & \cellcolor{lightlightlightyellow}60.8\%\textcolor{gray}{(+3.6\%)} & \cellcolor{lightlightlightblue}71.5\%\textcolor{gray}{(+1.5\%)}  \\ 

        \bottomrule\bottomrule
        \end{tabular}
    }
\end{center}
\vspace{-2.5em}
\end{table}

\begin{figure*}[t]
\centering
\includegraphics[width=0.99\linewidth]{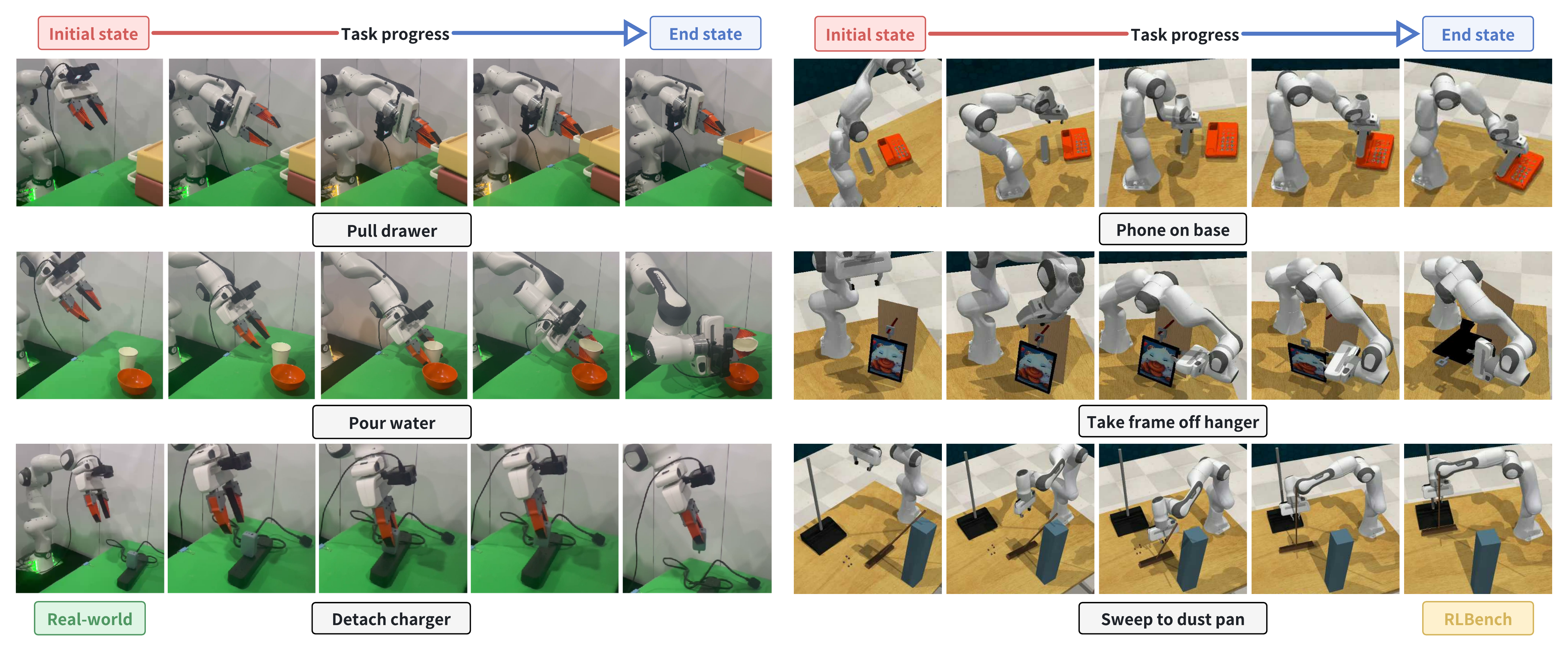}
\vspace{-0.2cm}
\caption{\textbf{The qualitative results of MoLe-VLA in both RLBench and real-world}, including the manipulation progress and the task completion end state for both simulation and real-world environments, are shown. More visualizations can be found in the Appendix.
}
\label{fig:realvis} 
\vspace{-1em}
\end{figure*}

\vspace{-1em}
\paragraph{Efficiency analysis}
To demonstrate the efficiency of MoLe-VLA, we analyze success rate changes with increasing skipped layers in \cref{fig:efficiency}. MoLe achieves similar success rates compared to the full-layer backbone while only \textit{19\%} of the FLOPs and delivering $\times 2$ faster inference. Notably, MoLe-OpenVLA significantly outperforms the original OpenVLA by a large margin.
Furthermore, detailed statistics on model efficiency are provided in \cref{tab:efficiency}. MoLe achieves the highest efficiency, requiring only \textit{0.309 seconds} per iteration during inference while maintaining the highest mean success rate of \textit{60.8\%}. These results highlight the superiority of MoLe in balancing efficiency and performance.

\vspace{-1em}
\paragraph{MoLe with quantization analysis}
We highlight the efficiency of MoLe under 8-bit quantization, which is more representative of real-world deployment scenarios, compared to FP16 CogAct as shown in ~\cref{tab:quant} on a commercial-grade RTX 4090D. MoLe achieves a higher success rate of 58.8\% with an inference frequency of 15.7 Hz, while utilizing only 55\% of the GPU memory compared to CogAct, which achieves just 9.8 Hz. This demonstrates MoLe's ability to maintain superior performance with significantly lower computational costs after quantization.

\begin{table}[t]
\begin{center}
    \setlength\tabcolsep{4pt} 
    \caption{\label{tab:ablation} \textbf{Ablation study} on STAR Router and CogKD loss with its variants on RLBench simulation environment.}
    \resizebox{0.99\columnwidth}{!}{%
        \begin{tabular}{ccccccc}
        \toprule\toprule
        \textbf{Methods} & \cellcolor{lightgray}STAR & \cellcolor{lightgray}Cognition &\multicolumn{3}{c}{\cellcolor{lightgray}CogKD Loss} & Mean$\uparrow$ \\
        \cmidrule(lr){4-6} 
        \rowcolor{lightblue}\textbf{\textit{RLBench}} & & & MSE & KL & Reserve KL & \\
        \midrule\midrule
        \rowcolor{lightlightlightred}
        \multicolumn{1}{c}{\cellcolor{lightred}$Ex_{0}$}   & \textcolor{red}{\ding{55}} & \textcolor{red}{\ding{55}} & \textcolor{red}{\ding{55}} & \textcolor{red}{\ding{55}} & \textcolor{red}{\ding{55}} & 57.2\% \\ 
        \rowcolor{lightlightlightyellow}
        \multicolumn{1}{c}{\cellcolor{lightyellow}$Ex_{1-1}$} & \textcolor{green}{\ding{51}} & \textcolor{red}{\ding{55}} & \textcolor{red}{\ding{55}} & \textcolor{red}{\ding{55}} & \textcolor{red}{\ding{55}} & 56.3\% \\ 
        \rowcolor{lightlightyellow}
        \multicolumn{1}{c}{\cellcolor{lightyellow}$Ex_{1-2}$} & \textcolor{green}{\ding{51}} & \textcolor{red}{\ding{55}} & \textcolor{green}{\ding{51}} & \textcolor{red}{\ding{55}} & \textcolor{red}{\ding{55}} & 54.8\% \\ 
        \multicolumn{1}{c}{\cellcolor{lightgreen}$Ex_{2-1}$} & \textcolor{green}{\ding{51}} & \textcolor{green}{\ding{51}} & \textcolor{green}{\ding{51}} & \textcolor{red}{\ding{55}} & \textcolor{red}{\ding{55}} & 58.3\% \\
        \rowcolor{lightlightlightgreen}
        \multicolumn{1}{c}{\cellcolor{lightgreen}$Ex_{2-2}$} & \textcolor{green}{\ding{51}} & \textcolor{green}{\ding{51}} & \textcolor{red}{\ding{55}} & \textcolor{green}{\ding{51}} & \textcolor{red}{\ding{55}} & 57.7\% \\
        \rowcolor{lightlightgreen}
        \multicolumn{1}{c}{\cellcolor{lightgreen}$Ex_{2-3}$} & \textcolor{green}{\ding{51}} & \textcolor{green}{\ding{51}} & \textcolor{red}{\ding{55}} & \textcolor{red}{\ding{55}} & \textcolor{green}{\ding{51}} & 59.4\% \\
        \rowcolor{lightgreen}
        \multicolumn{1}{c}{\cellcolor{lightgreen}$Ex_{2-4}$} & \textcolor{green}{\ding{51}} & \textcolor{green}{\ding{51}} & \textcolor{green}{\ding{51}} & \textcolor{red}{\ding{55}} & \textcolor{green}{\ding{51}} & \textbf{60.8\%} \\
        \bottomrule\bottomrule
        \end{tabular}
    }
\end{center}
\vspace{-2em}
\end{table}

\vspace{-1em}
\paragraph{Scalability evaluation} 
Table \ref{tab:scalability} highlights the scalability of our proposed MoLe compared to full-layer CogAct across different model sizes evaluated on the RLBench. MoLe consistently achieves higher mean success rates, with improvements of \textit{+2.7\%}, \textit{+3.6\%}, and \textit{+1.5\%} for Small, Base, and Large models, respectively. Notably, MoLe-Large achieves a mean success rate of \textit{71.5\%}, demonstrating its ability to leverage increased model capacity effectively. These results validate the robustness and adaptability of MoLe across diverse computational budgets and model scales.

\vspace{-1em}
\paragraph{Ablation study}
\label{sec:ablation}
Table~\ref{tab:ablation} demonstrates the effectiveness of our \textit{STAR} and \textit{CogKD} in the RLBench simulation environment. The baseline CogAct ($Ex_{0}$) achieves a mean success rate of \textit{57.2\%}, while integrating STAR with cognition tokens ($Ex_{2-1}$) boosts performance to \textit{58.3\%}, showcasing their synergy. Further improvements are observed with tailored CogKD loss variants, where combining STAR, cognition tokens, and Reserve KL loss ($Ex_{2-3}$) achieves \textit{59.4\%}, and the best performance of \textit{60.8\%} is achieved by adding both MSE and Reserve KL losses ($Ex_{2-4}$), a \textit{+3.6\%} gain over the baseline. These results highlight the strength of STAR in capturing spatial-temporal dependencies and the importance of cognition tokens for self-knowledge distillation.

\subsection{Evaluation with real-world tasks.} 
\label{sec:realworld}
We conducted experiments involving interactions with various real-world objects, as summarized in \cref{tab:real}. The results show that MoLe consistently delivers strong performance across three tasks. Notably, in the challenging \textit{pour water} task, which demands precise 3D position and rotation predictions, MoLe achieved an impressive success rate of \textit{80\%}. These results highlight that MoLe preserves the ability to understand 3D spatial scenes and make accurate predictions with a \textit{50\%} reduction in LLM computational cost.

\begin{table}[t]
\begin{center}
    \setlength\tabcolsep{12pt} 
    \caption{\label{tab:real} \textbf{Success rate for real-world} evaluated on the FR3 robot equipped with a 3D-preinted UMI gripper.}
    \resizebox{0.99\columnwidth}{!}{%
        \begin{tabular}{ccccc}
        \toprule\toprule
        \multirow{2}{*}{\textbf{Methods}} & \cellcolor{lightred}\textbf{Detach} & \cellcolor{lightgreen}\textbf{Pull} & \cellcolor{lightblue}\textbf{Pour} &
        \multirow{2}{*}{\textbf{Mean$\uparrow$}}\\
        & \cellcolor{lightred}\textbf{charger} &  \cellcolor{lightgreen}\textbf{drawer} & \cellcolor{lightblue}\textbf{water} & 
         \\
        \midrule\midrule

        \cellcolor{lightlightlightgray}MoLe   & \cellcolor{lightlightlightred}70.0\% & \cellcolor{lightlightlightgreen}60.0\% & \cellcolor{lightlightlightblue}80.0\% & 
        70.0\% \\ 
        \cellcolor{lightlightlightgray}CogAct   & \cellcolor{lightlightlightred}60.0\% & \cellcolor{lightlightlightgreen}60.0\% & \cellcolor{lightlightlightblue}80.0\% & 
        66.7\% \\ 

        \bottomrule\bottomrule
        \end{tabular}
    }
\end{center}
\vspace{-2em}
\end{table}

\subsection{Qualitative results}
\label{sec:qualitative}
As shown in \cref{fig:realvis}, we visualize the manipulation process for three real-world and three RLBench simulation tasks. Our method accurately predicts continuous 7-DoF end-effector poses, enabling precise task execution along planned trajectories. For instance, in the \textit{pour water} task, MoLe-VLA successfully grasps the cup, lifts the can, positions it above the bowl, and smoothly rotates the gripper to control water flow. Detailed demonstrations are provided in the supplementary video, with failure cases analyzed in the appendix.

\section{Conclusion}
\label{sec:conclusion}
We proposed \textbf{\textit{MoLe-VLA}}, a framework inspired by the Shallow Brain Hypothesis, to optimize VLA models for robotics. MoLe dynamically activates key LLM layers with a specially devised STAR router, reducing redundancy while preserving essential information. To address performance loss from layer skipping, we developed CogKD to enhance efficiency and cognitive capacity. Experiments on real-world and RLBench environments show that MoLe reduces computational costs, enabling efficient and adaptable robotic systems.

\section{Acknowledgment}
This work was supported by the National Natural Science Foundation of China (62476011).
{
    \small
    \bibliographystyle{ieeenat_fullname}
    \bibliography{main}
}
\clearpage
\appendix
\clearpage
\setcounter{page}{1}
\maketitlesupplementary
\definecolor{lightgray}{gray}{.9}
\definecolor{lightblue}{RGB}{230,240,255}
\definecolor{lightgreen}{RGB}{230,255,230}
\definecolor{lightyellow}{RGB}{255,255,230}
\definecolor{lightred}{RGB}{255,230,230}

\definecolor{lightlightgray}{gray}{.95}
\definecolor{lightlightblue}{RGB}{240,245,255}
\definecolor{lightlightgreen}{RGB}{240,255,240}
\definecolor{lightlightyellow}{RGB}{255,255,240}
\definecolor{lightlightred}{RGB}{255,240,240}

\definecolor{lightlightlightgray}{gray}{.99}
\definecolor{lightlightlightblue}{RGB}{247,250,255}
\definecolor{lightlightlightgreen}{RGB}{247,255,247}
\definecolor{lightlightlightyellow}{RGB}{255,255,247}
\definecolor{lightlightlightred}{RGB}{255,247,247}

The supplementary materials accompanying this paper provide an extensive quantitative and qualitative analysis of the proposed method. First, we show the deployment of real-world robots in \cref{ap:franka}. Then, we present the complete set of hyperparameters used in our experiments, detailed in \cref{ap:hyper}, to ensure reproducibility and facilitate further exploration by the research community. In \cref{ap:small}, we examine the training scalability of our proposed method, highlighting its performance across varying scales of data and model configurations.
Furthermore, we provide complete experiment results on RLBench, exploring the impact of different hyperparameters, including $\lambda_1$, $\lambda_2$, and $\lambda_3$. These experiments follow the same settings described in the main manuscript, and the results are summarized in \cref{ap:factor}. We also investigate the impact of skipping different numbers of layers in the RLBench environment, providing a complete evaluation of the trade-offs between efficiency and performance, as detailed in \cref{ap:skip}.
In addition, we include further qualitative analyses in \cref{ap:qu}, offering visual and descriptive insights into our method's performance enhancements and capabilities. This section highlights specific examples where our approach excels, emphasizing its ability to effectively handle diverse scenarios and complex tasks. Finally, we analyze the failure cases of our proposed methods in real-world environments in \cref{ap:fail}. These supplementary materials aim to provide a deeper understanding of the proposed method, supporting its robustness and applicability to various tasks.

\section{Real-world Franka robot setup}
\label{ap:franka}
\begin{figure}[t]
\includegraphics[width=0.45\textwidth]{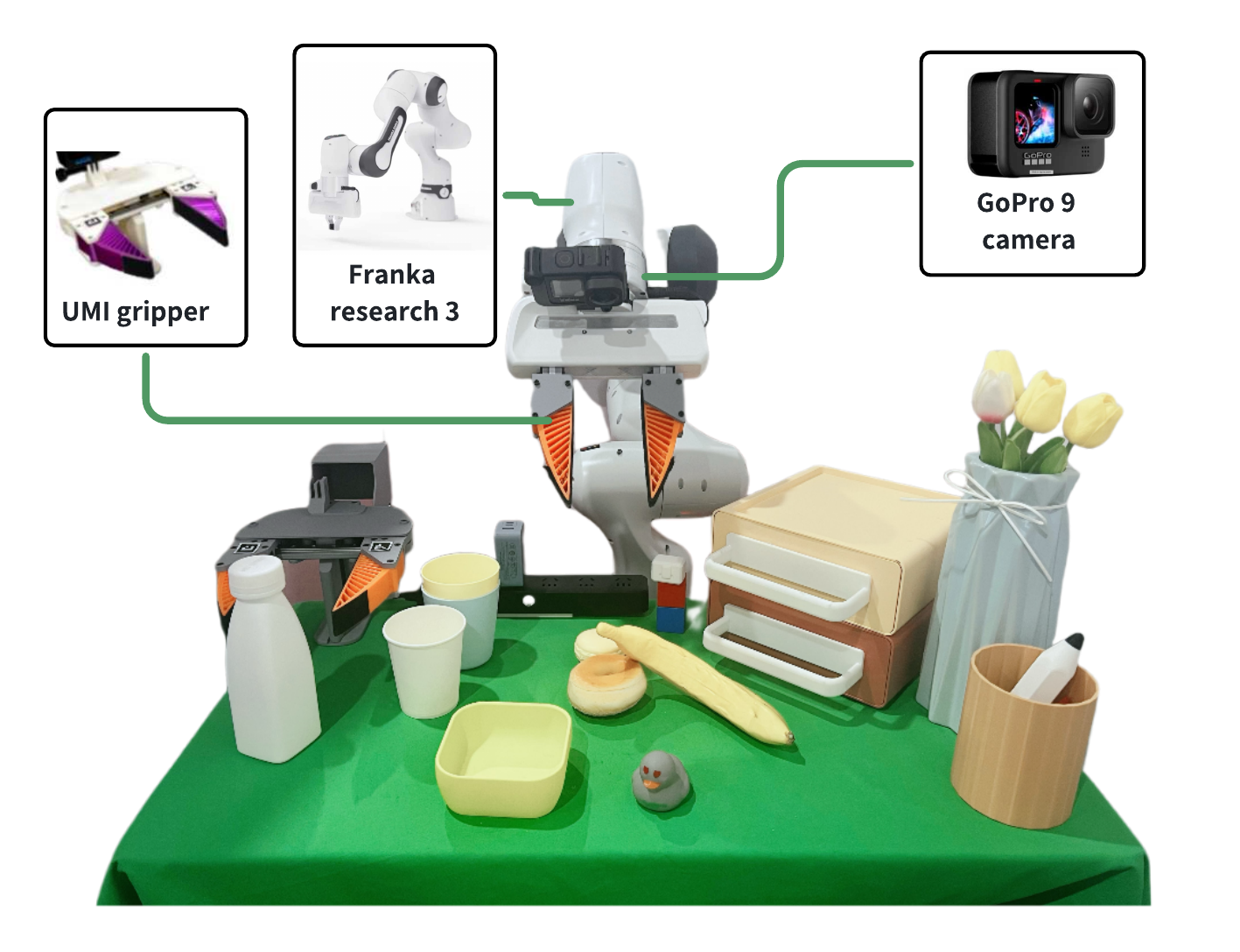}
\centering
\caption{Franka robot setup.}
\label{fig:franka}
\end{figure}

For our real-world experiments, we utilize the Franka Research 3 (FR3) robotic arm as the hardware platform. To overcome the limitations of the FR3's default gripper, which has relatively short fingers and struggles with certain complex tasks, we 3D-printed and replaced it with a UMI gripper~\cite{chi2024universal}. A GoPro 9 camera is positioned to the right of the setup to capture high-quality RGB images, providing visual input for the pipeline.
We conduct experiments on three tasks: \textit{detach charger}, \textit{pull drawer}, and \textit{pour water}. Keyframes are extracted to construct the training set for each task, with \textbf{10} frames used for each. Figure~\ref{fig:franka} illustrates the experimental setup and assets.
During the evaluation, task success is determined through human assessment. All actions are performed within the robot's coordinate system to ensure precision and consistency throughout the process. The successful outcomes of the three tasks are shown in the "End State" images in Figure 5 of the main text.

\begin{table}[t]
\small
\caption{Training hyper-parameters for RLBench.}
\label{tab:hyper}
\centering
\setlength{\tabcolsep}{25pt}
\begin{tabular}{cc}
\toprule
Hyper-parameters          & Values \\
\midrule
batch size                & 64*8                \\
optimizer                 & AdamW            \\
MLLM learning rate        & 2e-5             \\
action head learning rate & 2e-5           \\
learninrg rate schedule   & constant         \\
warmup steps              & 2500             \\
LSTM dropout              & 0.3              \\
MLP dropout               & 0.4              \\
training epochs                     & 100    \\
$\lambda$ & 0.05 \\
LSTM window size & 12 \\
\bottomrule
\end{tabular}
\end{table}

\newcolumntype{C}[1]{>{\columncolor{#1}}c}
\begin{table*}[t]
\caption{\textbf{Performance comparison with existing VLA models across ten tasks in RLBench settings.} We colour-coded the results \textbf{\textbf{red}} \textbf{(1st)} and \textcolor{blue}{blue} (2nd) and the row colour reflects the baseline type. The four efficiency methods operate with only 50\% LLM layers.} 
\label{tab:factor}
\resizebox{\linewidth}{!}{%
\setlength{\tabcolsep}{2mm}{
\begin{tabular}{ccccccc}
\toprule\toprule
\multirow{2}{*}{\textbf{Methods}}  & \multirow{2}{*}{Backbone} & \cellcolor{lightgray}Close  & \cellcolor{lightgray}Put Rubbish & \cellcolor{lightgray}Sweep & \cellcolor{lightgray}Phone & \cellcolor{lightgray}Change \\
 & & \cellcolor{lightgray}Fridge & \cellcolor{lightgray}in Bin & \cellcolor{lightgray}to Dustpan & \cellcolor{lightgray}on Base & \cellcolor{lightgray}Clock\\
\midrule\midrule
\rowcolor{lightblue}\multicolumn{7}{c}{\textbf{\textit{RLBench}}} \\ 
\midrule
\rowcolor{lightlightlightgreen}
\multicolumn{1}{c}{\cellcolor{lightgreen}$\lambda_1=0.5, \lambda_2=0.5, \lambda_3=0.5, \alpha=0.999$} & MoLe-CogAct & 80.0\% & 16.0\% & 60.0\% & 36.0\% & 24.0\%  \\
\rowcolor{lightlightlightgreen}
\multicolumn{1}{c}{\cellcolor{lightgreen}$\lambda_1=0.5, \lambda_2=0.5, \lambda_3=0.5, \alpha=0.9999$} & MoLe-CogAct & 84.0\% & 20.0\% & 52.0\% & 28.0\% & 12.0\%  \\
\rowcolor{lightlightlightyellow}
\multicolumn{1}{c}{\cellcolor{lightyellow}$\lambda_1=0.5, \lambda_2=0.8, \lambda_3=0.5, \alpha=0.999$} & MoLe-CogAct & 76.0\% & 24.0\% & 32.0\% & 24.0\% & 36.0\%  \\
\rowcolor{lightlightlightyellow}
\multicolumn{1}{c}{\cellcolor{lightyellow}$\lambda_1=0.5, \lambda_2=0.8, \lambda_3=0.5, \alpha=0.9999$} & MoLe-CogAct & 68.0\% & 8.0\% & 44.0\% & 40.0\% & 44.0\%  \\
\rowcolor{lightlightlightyellow}
\multicolumn{1}{c}{\cellcolor{lightyellow}$\lambda_1=0.5, \lambda_2=0.5, \lambda_3=0.1, \alpha=0.999$} & MoLe-CogAct & 60.0\% & 20.0\% & 60.0\% & 48.0\% & 28.0\%  \\
\rowcolor{lightlightlightyellow}
\multicolumn{1}{c}{\cellcolor{lightyellow}$\lambda_1=0.8, \lambda_2=0.5, \lambda_3=0.1, \alpha=0.999$}  & MoLe-CogAct & 96.0\% & 36.0\% & 8.0\% & 52.0\% & 16.0\%  \\
\rowcolor{lightlightlightred}
\multicolumn{1}{c}{\cellcolor{lightred}$\lambda_1=0.5, \lambda_2=0.5, \lambda_3=1.0, \alpha=0.999$} & MoLe-CogAct & 96.0\% & 64.0\% & 28.0\% & 52.0\% & 20.0\%  \\
\rowcolor{lightlightlightred}
\multicolumn{1}{c}{\cellcolor{lightred}$\lambda_1=0.5, \lambda_2=0.1, \lambda_3=0.5, \alpha=0.999$}  & MoLe-CogAct & 88.0\% & 24.0\% & 68.0\% & 36.0\% & 20.0\%  \\
\midrule
\multirow{2}{*}{\textbf{Methods}} & \cellcolor{lightgray}Take Umbrella & \cellcolor{lightgray}Take Frame & \cellcolor{lightgray}Close  & \cellcolor{lightgray}Close & \cellcolor{lightgray}Toilet & \multirow{2}{*}{Mean Acc.\% $\uparrow$}  \\
& \cellcolor{lightgray}out of Stand & \cellcolor{lightgray}off Hanger & \cellcolor{lightgray}Box & \cellcolor{lightgray}Laptop Lid & \cellcolor{lightgray}Seat Down & \\
\midrule

\rowcolor{lightlightlightgreen}
\multicolumn{1}{c}{\cellcolor{lightgreen}$\lambda_1=0.5, \lambda_2=0.5, \lambda_3=0.5, \alpha=0.999$} & 48.0\% & 72.0\% & 76.0\% & 68.0\% & 84.0\% & 53.8\%  \\
\rowcolor{lightlightlightgreen}
\multicolumn{1}{c}{\cellcolor{lightgreen}$\lambda_1=0.5, \lambda_2=0.5, \lambda_3=0.5, \alpha=0.9999$} & 52.0\% & 68.0\% & 80.0\% & 56.0\% & 80.0\% & 53.2\%  \\
\rowcolor{lightlightlightyellow}
\multicolumn{1}{c}{\cellcolor{lightyellow}$\lambda_1=0.5, \lambda_2=0.8, \lambda_3=0.5, \alpha=0.999$} & 60.0\% & 64.0\% & 76.0\% & 60.0\% & 92.0\% & 54.4\%  \\
\rowcolor{lightlightlightyellow}
\multicolumn{1}{c}{\cellcolor{lightyellow}$\lambda_1=0.5, \lambda_2=0.8, \lambda_3=0.5, \alpha=0.9999$} & 48.0\% & 68.0\% & 80.0\% & 56.0\% & 84.0\% & 54.0\%  \\
\rowcolor{lightlightlightyellow}
\multicolumn{1}{c}{\cellcolor{lightyellow}$\lambda_1=0.5, \lambda_2=0.5, \lambda_3=0.1, \alpha=0.999$} & 40.0\% & 60.0\% & 96.0\% & 68.0\% & 92.0\% & 57.2\%  \\
\rowcolor{lightlightlightyellow}
\multicolumn{1}{c}{\cellcolor{lightyellow}$\lambda_1=0.8, \lambda_2=0.5, \lambda_3=0.1, \alpha=0.999$}  & 52.0\% & 56.0\% & 84.0\% & 60.0\% & 100.0\% & 56.0\%  \\
\rowcolor{lightlightlightred}
\multicolumn{1}{c}{\cellcolor{lightred}$\lambda_1=0.5, \lambda_2=0.5, \lambda_3=1.0, \alpha=0.999$} & 68.0\% & 24.0\% & 80.0\% & 72.0\% & 92.0\% & 59.6\%  \\
\rowcolor{lightlightlightred}
\multicolumn{1}{c}{\cellcolor{lightred}$\lambda_1=0.5, \lambda_2=0.1, \lambda_3=0.5, \alpha=0.999$}  & 36.0\% & 72.0\% & 84.0\% & 80.0\% & 100.0\% & 60.8\%  \\
\bottomrule\bottomrule
\end{tabular}%
}}
\end{table*}

\section{Training details}
\label{ap:hyper}
We conducted experiments on the RLBench benchmark using the hyperparameters summarized in Table~\ref{tab:hyper}. The model was trained with a batch size of $64 \times 8$, and the AdamW optimizer was utilized for optimization. The learning rate for the MLLM was set to $2 \times 10^{-5}$, while the action head learning rate was configured as $2 \times 10^{-5}$. A constant learning rate schedule was adopted, with 2500 warmup steps to stabilize training at the initial stages.
To prevent overfitting and enhance generalization, we applied a dropout rate of 0.3 for the LSTM layers and 0.4 for the MLP layers. The LSTM window size was configured to 12 to effectively capture temporal dependencies in sequential data. The training process is set to 100 epochs. Additionally, we set the regularization parameter $\lambda$ to 0.05 to balance different loss components.
This setup was chosen to ensure stable and efficient training while maximizing performance on the RLBench tasks. These hyperparameters were fine-tuned based on preliminary experiments to achieve optimal results.

\section{Data scalability}
\label{ap:small}
To evaluate the data scalability of our MoLe model, we conducted experiments on a reduced dataset comprising only three tasks: \textit{Close box}, \textit{Close laptop lid}, and \textit{Toilet seat down}. The results, summarized in \cref{tab:small}, demonstrate that MoLe consistently achieves the highest success rate with \textit{82.7\%} across all tasks, outperforming other methods such as CogAct with \textit{71.0\%}, Random-skip with \textit{64.1\%}, and DeeR with \textit{78.6\%}. Notably, despite being trained on fewer tasks, MoLe maintains superior performance while utilizing only \textit{50\%} of the computational resources compared to the baseline models. These findings highlight the strong data scalability and computational efficiency of MoLe, making it particularly effective in scenarios with limited training data or constrained computational budgets.

\begin{table}[t]
\begin{center}
    \setlength\tabcolsep{4pt} 
    \caption{\label{tab:small} \textbf{Data scalability analysis} evaluated on RLBench simulationn environment with only three tasks.}
    \resizebox{0.99\columnwidth}{!}{%
        \begin{tabular}{ccccc}
        \toprule\toprule
        \textbf{Methods} & \cellcolor{lightgray}\textbf{Close box$\downarrow$} & \cellcolor{lightgray}\textbf{Close laptop$\downarrow$} & \cellcolor{lightgray}\textbf{Seat down$\downarrow$} & \textbf{Mean$\uparrow$} \\
        \midrule\midrule

        \cellcolor{lightblue}CogAct   & \cellcolor{lightlightlightblue}96.0\% & \cellcolor{lightlightlightblue}84.0\%  & \cellcolor{lightlightlightblue}32.0\%  & \cellcolor{lightlightlightblue}71.0\% \\ 

        \cellcolor{lightyellow}Random-skip   & \cellcolor{lightlightlightyellow}84.0\% & \cellcolor{lightlightlightyellow}60.0\% & \cellcolor{lightlightlightyellow}52.0\% & \cellcolor{lightlightlightyellow}64.1\% \\ 
        
        \cellcolor{lightgreen}DeeR    & \cellcolor{lightlightlightgreen}100.0\% & \cellcolor{lightlightlightgreen}76.0\% & \cellcolor{lightlightlightgreen}60.0\%  & \cellcolor{lightlightlightgreen}78.6\%  \\ 
                
        \cellcolor{lightred}MoLe   & \cellcolor{lightlightlightred}100.0\% & \cellcolor{lightlightlightred}84.0\% & \cellcolor{lightlightlightred}64.0\% & \cellcolor{lightlightlightred}82.7\%  \\ 

        \bottomrule\bottomrule
        \end{tabular}
    }
\end{center}
\vspace{-2em}
\end{table}

\newcolumntype{C}[1]{>{\columncolor{#1}}c}
\begin{table*}[t]
\caption{\textbf{Performance comparison with existing VLA models across ten tasks in RLBench settings.} We colour-coded the results \textbf{\textbf{red}} \textbf{(1st)} and \textcolor{blue}{blue} (2nd) and the row colour reflects the baseline type. The four efficiency methods operate with only 50\% LLM layers.} 
\label{tab:skip}
\resizebox{\linewidth}{!}{%
\setlength{\tabcolsep}{6mm}{
\begin{tabular}{ccccccc}
\toprule\toprule
\multirow{2}{*}{\textbf{Methods}}  & \multirow{2}{*}{Backbone} & \cellcolor{lightgray}Close  & \cellcolor{lightgray}Put Rubbish & \cellcolor{lightgray}Sweep & \cellcolor{lightgray}Phone & \cellcolor{lightgray}Change \\
 & & \cellcolor{lightgray}Fridge & \cellcolor{lightgray}in Bin & \cellcolor{lightgray}to Dustpan & \cellcolor{lightgray}on Base & \cellcolor{lightgray}Clock\\
\midrule\midrule
\rowcolor{lightblue}\multicolumn{7}{c}{\textbf{\textit{RLBench}}} \\ 
\midrule
\rowcolor{lightlightlightgreen}
\multicolumn{1}{c}{\cellcolor{lightgreen}Skip 2 layers} & MoLe-CogAct & 92.0\% & 48.0\% & 56.0\% & 56.0\% & 20.0\%  \\
\rowcolor{lightlightlightgreen}
\multicolumn{1}{c}{\cellcolor{lightgreen}Skip 6 layers} & MoLe-CogAct & 88.0\% & 52.0\% & 48.0\% & 60.0\% & 24.0\%  \\
\rowcolor{lightlightlightyellow}
\multicolumn{1}{c}{\cellcolor{lightyellow}Skip 8 layers} & MoLe-CogAct & 96.0\% & 64.0\% & 36.0\% & 56.0\% & 56.0\%  \\
\rowcolor{lightlightlightyellow}
\multicolumn{1}{c}{\cellcolor{lightyellow}Skip 12 layers} & MoLe-CogAct & 100.0\% & 56.0\% & 40.0\% & 60.0\% & 48.0\%  \\
\rowcolor{lightlightlightyellow}
\multicolumn{1}{c}{\cellcolor{lightyellow}Skip 20 layers} & MoLe-CogAct & 80.0\% & 24.0\% & 36.0\% & 48.0\% & 16.0\%  \\
\rowcolor{lightlightlightyellow}
\multicolumn{1}{c}{\cellcolor{lightyellow}Skip 24 layers}  & MoLe-CogAct & 72.0\% & 40.0\% & 52.0\% & 40.0\% & 16.0\%  \\
\rowcolor{lightlightlightred}
\multicolumn{1}{c}{\cellcolor{lightred}Skip 26 layers} & MoLe-CogAct & 80.0\% & 44.0\% & 48.0\% & 36.0\% & 24.0\%  \\
\rowcolor{lightlightlightred}
\multicolumn{1}{c}{\cellcolor{lightred}Skip 30 layers}  & MoLe-CogAct & 56.0\% & 0.0\% & 0.0\% & 32.0\% & 40.0\%  \\
\midrule
\multirow{2}{*}{\textbf{Methods}} & \cellcolor{lightgray}Take Umbrella & \cellcolor{lightgray}Take Frame & \cellcolor{lightgray}Close  & \cellcolor{lightgray}Close & \cellcolor{lightgray}Toilet & \multirow{2}{*}{Mean Acc.\% $\uparrow$}  \\
& \cellcolor{lightgray}out of Stand & \cellcolor{lightgray}off Hanger & \cellcolor{lightgray}Box & \cellcolor{lightgray}Laptop Lid & \cellcolor{lightgray}Seat Down & \\
\midrule

\rowcolor{lightlightlightgreen}
\multicolumn{1}{c}{\cellcolor{lightgreen}Skip 2 layers} & 48.0\% & 76.0\% & 96.0\% & 68.0\% & 92.0\% & 65.2\%  \\
\rowcolor{lightlightlightgreen}
\multicolumn{1}{c}{\cellcolor{lightgreen}Skip 6 layers} & 52.0\% & 64.0\% & 92.0\% & 56.0\% & 96.0\% & 63.2\%  \\
\rowcolor{lightlightlightyellow}
\multicolumn{1}{c}{\cellcolor{lightyellow}Skip 8 layers} & 32.0\% & 44.0\% & 80.0\% & 68.0\% & 84.0\% & 61.6\%  \\
\rowcolor{lightlightlightyellow}
\multicolumn{1}{c}{\cellcolor{lightyellow}Skip 12 layers} & 52.0\% & 52.0\% & 60.0\% & 56.0\% & 100.0\% & 62.4\%  \\
\rowcolor{lightlightlightyellow}
\multicolumn{1}{c}{\cellcolor{lightyellow}Skip 20 layers} & 56.0\% & 68.0\% & 76.0\% & 48.0\% & 100.0\% & 55.2\%  \\
\rowcolor{lightlightlightyellow}
\multicolumn{1}{c}{\cellcolor{lightyellow}Skip 24 layers} & 52.0\% & 48.0\% & 80.0\% & 52.0\% & 80.0\% & 53.2\%   \\
\rowcolor{lightlightlightred}
\multicolumn{1}{c}{\cellcolor{lightred}Skip 26 layers} & 48.0\% & 56.0\% & 72.0\% & 48.0\% & 84.0\% & 54.0\%  \\
\rowcolor{lightlightlightred}
\multicolumn{1}{c}{\cellcolor{lightred}Skip 30 layers}  & 44.0\% & 40.0\% & 68.0\% & 28.0\% & 76.0\% & 38.4\%  \\
\bottomrule\bottomrule
\end{tabular}%
}}
\end{table*}

\section{Hyperparameter analysis}
\label{ap:factor}
We conducted additional experiments on RLBench with different combinations of the hyperparameters. Table \ref{tab:factor} demonstrates the results of our parameter ablation study, evaluating different configurations of MoLe-CogAct across ten RLBench tasks under a 50\% layer-skip setting. The analysis highlights the impact of key hyperparameters ($\lambda_1$, $\lambda_2$, $\lambda_3$, and $\alpha$) on task performance.  
The results show that increasing the weight of $\lambda_3$, which emphasizes the role of specific layers, consistently improves performance, while smaller $\alpha$ values lead to better optimization stability. Additionally, the influence of $\lambda_2$ varies depending on the task, indicating its role in balancing intermediate layer contributions. These findings underline the adaptability of MoLe-CogAct and its ability to achieve strong performance and efficiency through careful parameter tuning.





\section{Layer skip analysis}
\label{ap:skip}
We provide complete experiments on RLBench with different numbers of skipping layers from 2 to 30. As shown in ~\cref{tab:skip}, our proposed MoLe-CogAct demonstrates remarkable robustness and efficiency as the number of skipped LLM layers increases. Even with substantial layer skipping, the performance remains stable across most tasks, with only a slight decline in success rates up to 24 skipping layers. Notably, it is only when skipping 30 layers, resulting in an almost 95\% reduction in FLOPs, that a significant drop in performance is observed. This highlights the exceptional efficiency of our method, which maintains strong task success rates while drastically reducing computational costs. These results underscore the robustness and adaptability of MoLe, making it a highly effective solution for efficient embodied intelligence tasks.

\begin{figure*}[t]
\centering
\includegraphics[width=0.7\linewidth]{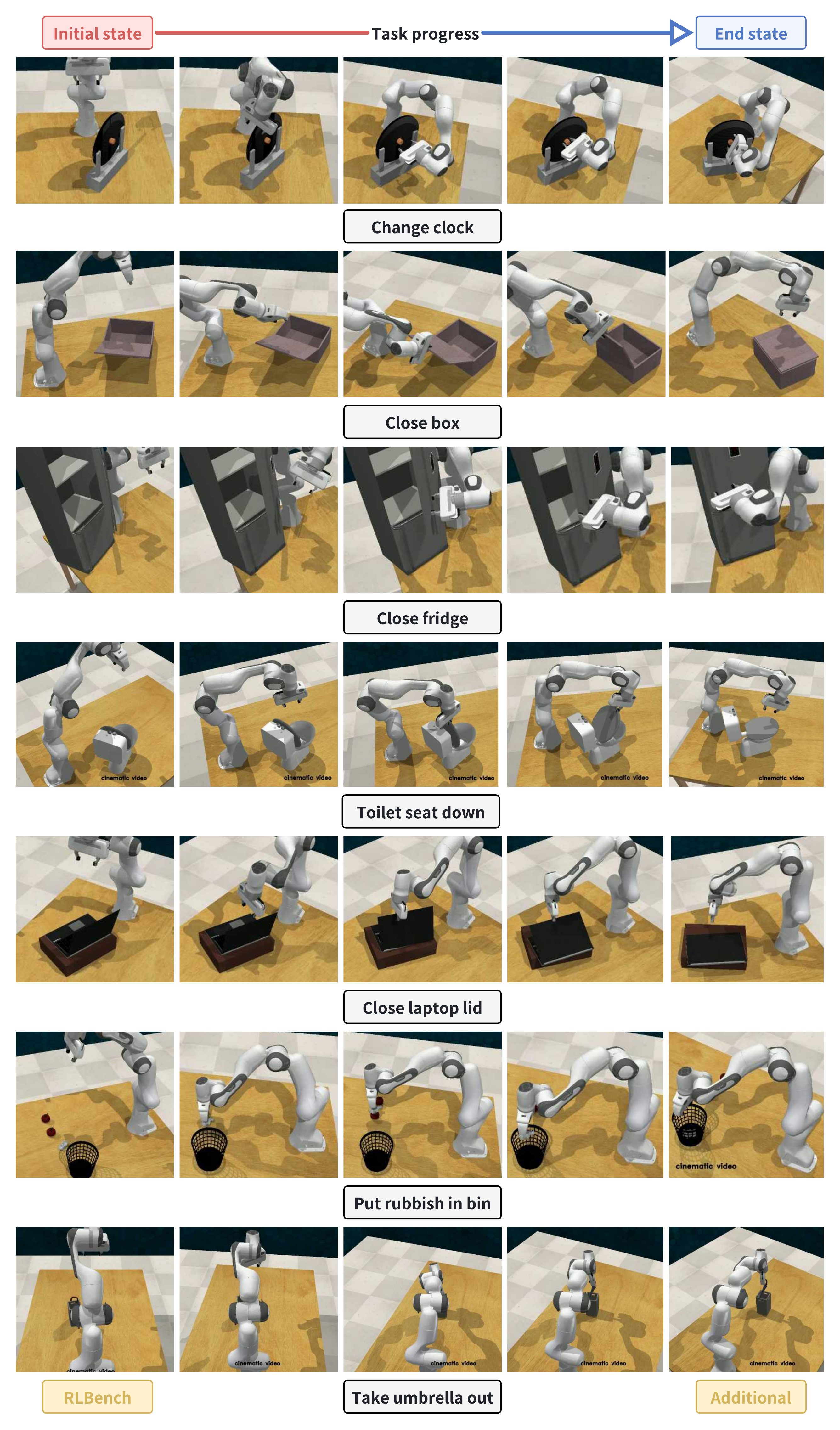}
\vspace{-0.2cm}
\caption{\textbf{The qualitative results of MoLe-VLA in RLBench simulation environment}, including the manipulation progress and the task completion end state.
}
\label{fig:apvis} 
\vspace{-0.2cm}
\end{figure*}

\section{Additional qualititive results}
\label{ap:qu}
This section presents visualizations of the manipulation processes for seven RLBench simulation tasks not covered in the main text. As shown in ~\cref{fig:apvis}, these visualizations illustrate task executions performed by our proposed MoLe-CogAct model. Each task is specifically designed to evaluate different capabilities of the efficient layer-skipping architecture.
Our method accurately predicts 7-DoF end-effector poses, enabling smooth and precise task completion along the defined trajectories. For example, in the \textit{put rubbish in bin} task, MoLe demonstrates a robust ability to grasp the rubbish accurately, lift it smoothly, and drop it precisely into the bin. This task exemplifies the model's spatial reasoning capabilities, requiring precise perception of both the rubbish and bin positions, as well as the ability to distinguish the rubbish from other objects in the scene.  
These results highlight MoLe's effectiveness in solving tasks that demand both spatial understanding and precise control. Demonstration videos of these tasks are provided in the supplementary material for further reference.

\section{Failure case analysis}
\label{ap:fail}
\begin{figure*}[t]
\centering
\includegraphics[width=0.7\linewidth]{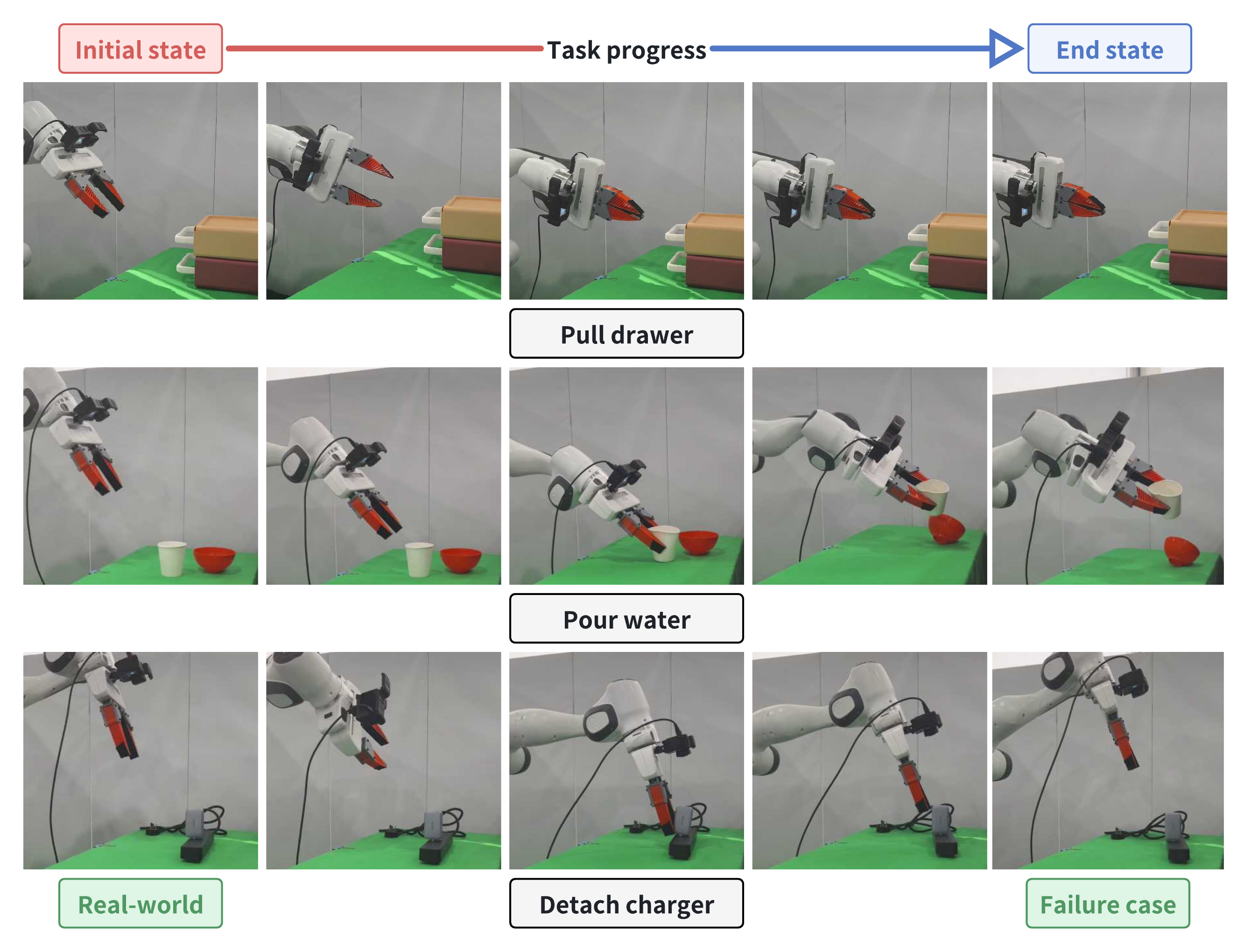}
\vspace{-0.2cm}
\caption{\textbf{The failure case analysis of MoLe-VLA in real-world environment}, including the manipulation progress and the task completion end state.
}
\label{fig:fail} 
\vspace{-0.2cm}
\end{figure*}

As shown in \cref{fig:fail}, through comprehensive real-world testing, we identified four key categories of failure cases that hinder MoLe's performance.
The first category is \textbf{loss of control}, which often occurs during interactions with target objects, such as in \textit{pull drawer}. These failures are marked by improper force application when handling objects of different weights or by the gripper slipping unexpectedly on smooth surfaces.
The second category involves \textbf{rotational prediction errors}, which are most evident in tasks requiring precise rotational control, such as \textit{pour water}. Failures in this group include incorrect angles during object interactions and cumulative errors in multi-step rotational motions.
The third category pertains to \textbf{pose predictions that exceed the robot's physical limits}. Here, the model occasionally predicts poses beyond the mechanical capabilities of the Franka robotic arm or generates unreachable target positions due to workspace constraints, as observed in tasks like \textit{detach charger}.
These failure cases suggest that while our method achieves significant efficiency gains through LLM layer skipping, it comes at the cost of reduced expressiveness and reasoning capacity in the LLM, particularly for tasks requiring fine-grained control and precise spatial understanding.

\end{document}